\documentclass{article}
\usepackage{ijcai22}

\usepackage{times}
\usepackage{soul}
\usepackage{url}
\usepackage[hidelinks]{hyperref}
\usepackage[utf8]{inputenc}
\usepackage[small]{caption}
\usepackage{graphicx}
\usepackage{amsmath}
\usepackage{amsthm}
\usepackage{booktabs}
\usepackage{algorithm}
\usepackage{algorithmic}

\usepackage{subfigure}
\usepackage{amsmath}
\usepackage{amsfonts}
\usepackage{multirow,makecell}
\usepackage[page]{appendix} % print appendices title
\title{Spatiality-guided Transformer for 3D Dense Captioning on Point Clouds}

\author{
Heng Wang
\and
Chaoyi Zhang\and
Jianhui Yu\And
Weidong Cai
\affiliations
School of Computer Science, University of Sydney, Australia\\
\emails
\{heng.wang, chaoyi.zhang, jianhui.yu, tom.cai\}@sydney.edu.au}
\begin{document}

\maketitle

\begin{abstract}
Dense captioning in 3D point clouds is an emerging vision-and-language task involving object-level 3D scene understanding. Apart from coarse semantic class prediction and bounding box regression as in traditional 3D object detection, 3D dense captioning aims at producing a further and finer instance-level label of natural language description on visual appearance and spatial relations for each scene object of interest. To detect and describe objects in a scene, following the spirit of neural machine translation, we propose a transformer-based encoder-decoder architecture, namely SpaCap3D, to transform objects into descriptions, where we especially investigate the relative spatiality of objects in 3D scenes and design a spatiality-guided encoder via a token-to-token spatial relation learning objective and an object-centric decoder for precise and spatiality-enhanced object caption generation. Evaluated on two benchmark datasets, ScanRefer and ReferIt3D, our proposed SpaCap3D outperforms the baseline method Scan2Cap by 4.94\% and 9.61\% in CIDEr@0.5IoU, respectively. Our project page with source code and supplementary files is available at \url{https://SpaCap3D.github.io/}.
\end{abstract}
% \footnote{The code will be released soon.}

\section{Introduction}
With continuous advance of deep learning in both computer vision and natural language processing, a variety of multimodal studies in these two areas have gained increasingly active attention~\cite{multimodal}. Dense captioning, as first introduced in image domain~\cite{densecap}, is a representative task among them to describe every salient pixel-formed area with a sequence of words. Just as many other multimodal tasks, the scope of conventional dense captioning research is mainly restricted to 2D space~\cite{2ddc1,2ddc2,2ddc3,2ddc4}. In recent past, with the popularity of 3D point-based scene data collection and application, 3D scene understanding and analysis have become feasible and prominent~\cite{qi2018frustum,scenegraph,scenegraph2,3dvg}. Also, two newly introduced dense annotation datasets tailored for 3D indoor scenes~\cite{scannet}, ScanRefer~\cite{scanrefer} and ReferIt3D~\cite{referit3d}, create more opportunities in 3D multimodal research. Facilitated by these datasets and pioneering point cloud processing techniques~\cite{pointnet++,votenet}, dense captioning has been recently lifted from 2D to 3D~\cite{scan2cap} to localize and describe each object in a 3D point cloud scene, which is beneficial for applications such as robotics manipulation, augmented reality, and autonomous driving.

% -------------- figure:newteaser ---------------
\begin{figure}[!t]
\centering     %%% not \center
\includegraphics[width=\linewidth]{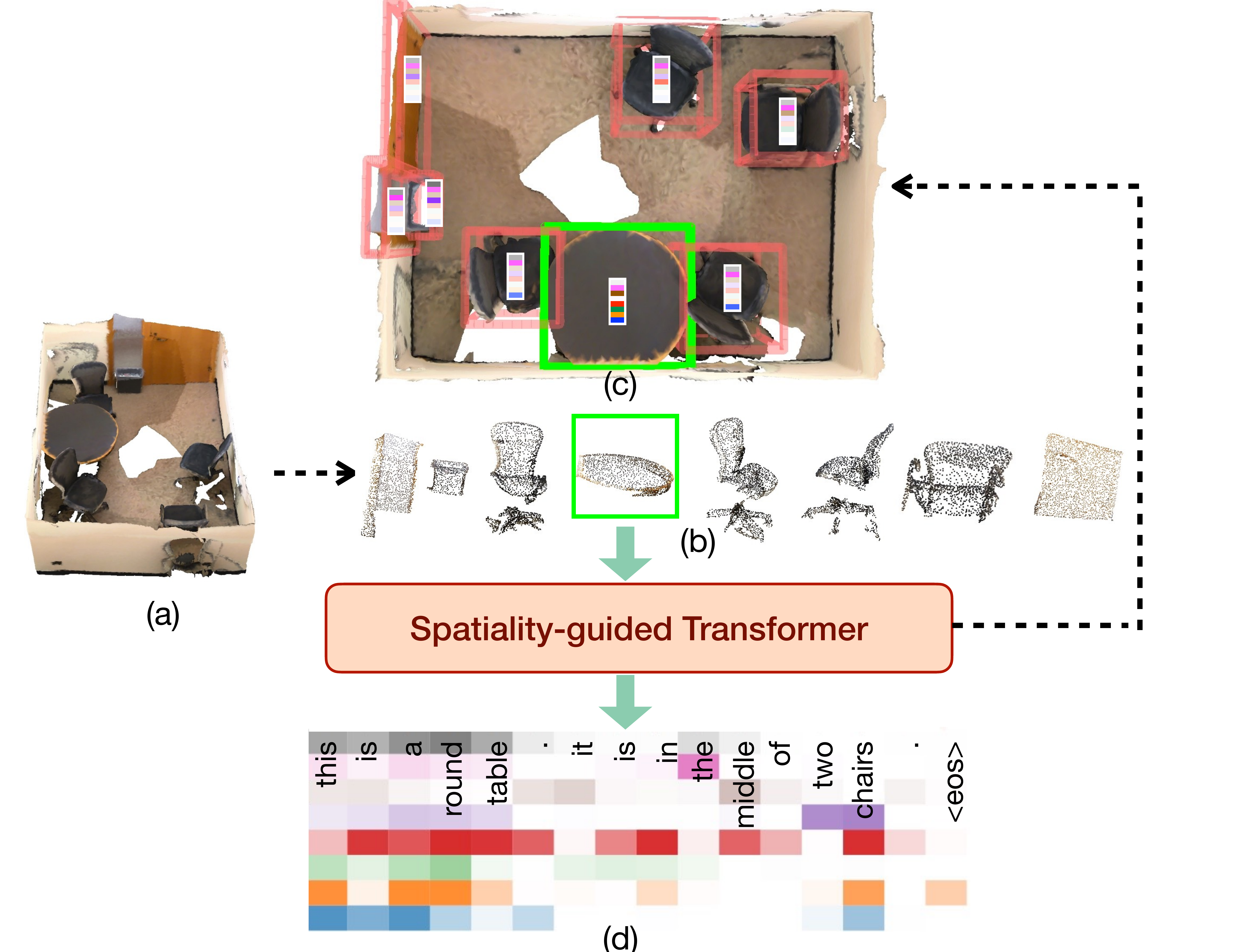}
\caption{Dense captioning for a target point-cloud object. Target and its surrounding objects are marked in green and red, respectively. (a) Point-based scene input. (b) Detected vision tokens.  (c) Neighbor-to-target contribution visualization in our encoder. (d) Target-to-word contribution visualization in our decoder. Detailed explanations including the color scheme used for attention heads, can be found in the supplementary.}
\label{fig:newteaser}
\end{figure}
% -------------- figure ---------------

In real world, human descriptions of an object or instructions to navigate a robot always involve good understanding and capturing of relative spatiality in 3D space~\cite{spatial1,spatial2}. In 3D dense captioning datasets, spatial language (\textit{above}, \textit{under}, \textit{left}, \textit{right}, \textit{in front of}, \textit{behind}, etc.) could be ubiquitous, taking up 98.7\% and 90.5\% in ScanRefer and ReferIt3D respectively according to their dataset statistics. However, such critical 3D spatiality has not been well explored in previous work~\cite{scan2cap}. Also, the sequential training strategy in their adopted RNN-based captioner could make it prohibitively long to reach convergence. In contrast, attention mechanism in prevalent Transformer~\cite{transformer} is capable of not only long-range relationship learning but also efficient parallel training. However, relation learning in transformer-based architectures depends only on the final task objective and lacks an explicit guidance on the relation, which could make it hard to precisely learn how 3D spatially-related an object is with respect to another one in our task.

To bridge the gap, in this work, we conduct careful \textit{relative spatiality modeling} to represent 3D spatial relations and propose a spatiality-guided Transformer for 3D dense captioning. Building upon a detection backbone which decomposes the input 3D scene into a set of object candidates (i.e., tokens), we propose \textbf{SpaCap3D}, a transformer-based encoder-decoder architecture, which consists of a spatiality-guided encoder where the relation learning among tokens is additionally supervised with a \textit{token-to-token spatial relationship} guidance whose labels are on-the-fly generated main-axis spatial relations  based on our relative spatiality modeling and an \textit{object-centric} decoder to transform each spatiality-enhanced vision token into a description, as shown in Figure~\ref{fig:newteaser}.  With faster training and efficient usage of data at hand, our proposed method exceeds Scan2Cap~\cite{scan2cap} by 4.94\% and 9.61\% in CIDEr@0.5IoU on ScanRefer~\cite{scanrefer} and Nr3D from ReferIt3D~\cite{referit3d}, respectively. To iterate, our contributions are three-fold: 
\begin{itemize}
\item We propose a token-to-token spatial relation learning objective with relative spatiality modeling to guide the encoding of main-axis spatial relations for better representation of 3D scene objects.
\item An integrated and efficient transformer-based encoder-decoder architecture, SpaCap3D, is proposed for 3D dense captioning, consisting of a spatiality-guided encoder and an object-centric decoder.
\item We achieve a new state-of-the-art performance on ScanRefer and Nr3D from ReferIt3D over previous work.
\end{itemize}

\section{Related Work}

\subsection{Dense Captioning: from 2D to 3D}
Following ~\cite{densecap}, current state-of-the-art 2D dense captioning methods use a region proposal network to detect salient regions and extract their CNN-based features as representation where a RNN captioner is applied to generate phrases or sentences. ~\cite{2ddc1} attached a late-fusion context feature extractor LSTM with a captioner one to emphasize contextual cues. ~\cite{2ddc2} and ~\cite{2ddc3} proposed to consider not only the global context but also the neighboring and the target-guided object context, respectively. In ~\cite{2ddc4}, a sub-pred-obj relationship was learnt via a triple-stream network. 

As 2D image is a projection of 3D world without depth dimension, spatial relations expressed in 2D dense captioning are usually implicit and ambiguous. To directly tackle 3D world, Scan2Cap~\cite{scan2cap} proposed 3D dense captioning on point cloud data. In Scan2Cap, relations among object proposals are learnt through a message passing network where only angular deviation relations whose labels~\cite{scan2cad} (i.e., transformation matrices) are hard to collect and incomplete are taken into consideration, while captions are generated by RNN-based decoder following 2D dense captioning methods, which is time-consuming in training. Compared to Scan2Cap, our work focuses on more common spatial relations, and the relation labels are easy to obtain for all objects during training as our label generation process only requires access to the bounding box information (i.e., box center and size). In addition, fast parallel-training in transformer-based architectures guarantees the efficiency of our method.

\subsection{Transformers in Image Captioning}
% 2d image classification, 2d object detection, 2d image captioning
% 3d visual grounding, 3d object detection
Although Transformers in dense captioning have not been explored to the best of our knowledge, there are a string of works in the related image captioning area. To learn better region representations, encoders in existing work were incorporated with learnable prior knowledge~\cite{m2}, geometric weight learnt from geometry features~\cite{relationtransformer}, region and attribute representations~\cite{entangle}, inter- and intra-layer global representations~\cite{ji2021improving}, or proposal- and grid-level features~\cite{luo2021dual}. Their decoders focus on how to learn the implicit relationship among region proposals so that a general and overall image-level caption can be generated. Yet, dense captioning in 3D world involves more diversities and degrees of freedom in object arrangements and it emphasizes finer and denser object-level descriptions, which captures more interactions between an object and its surrounding environment. To tackle these challenges, we use location-aware positional encoding to encode global position and a spatiality-guided encoder with token-to-token spatial relation learning objective to learn relative 3D spatial structures, while our object-centric decoder transforms each spatiality-enhanced object visual representation into a description.    

\section{Method}
% -------------- figure:overview ---------------
\begin{figure*}[!t]
\centering     %%% not \center
\includegraphics[width=\textwidth]{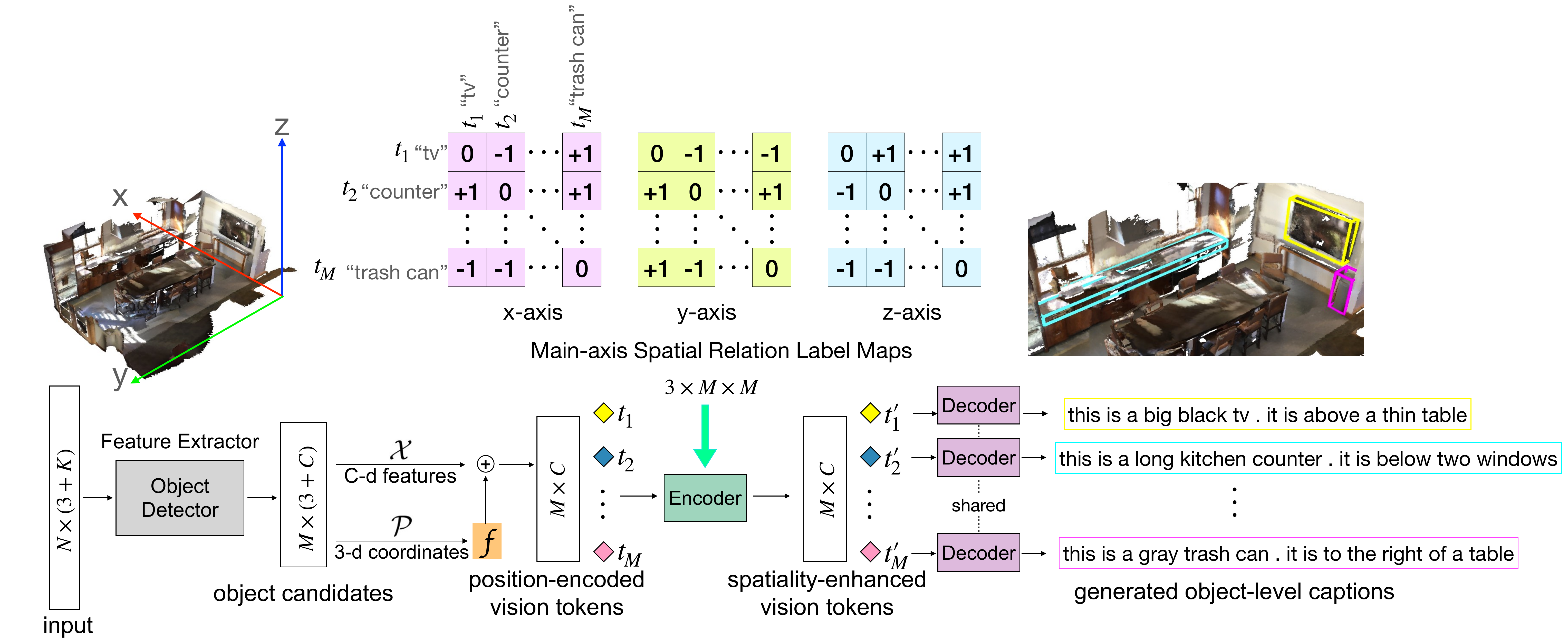}
\caption{The overview of our proposed method SpaCap3D for spatiality-guided 3D dense captioning. The encoder-decoder framework consists of an object detector to generate object proposals (i.e., tokens), a learnable function $f$ to project coordinates, a token-to-token spatial relation-guided encoder to incorporate relative 3D spatiality into tokens, and a shared object-centric decoder to generate per-object descriptions.}
\label{fig:overview}
\end{figure*}
% -------------- figure ---------------
We present our spatiality-guided Transformer as SpaCap3D for 3D dense captioning in Figure~\ref{fig:overview}. We first use a detector to decompose input scene into object proposals which we refer to as vision tokens, and then feed them into a spatiality-guided encoder for token-to-token relative 3D spatiality learning. Lastly, a shared object-centric decoder is conditioned on each spatiality-enhanced object vision token to describe them individually.

% \subsection{Revisit Transformers}
% Standard Transformer~\cite{transformer} consists of an encoder to encode input sentence into high-dimensional space and a decoder to decode such representation into a sentence in another language to achieve machine translation. Each encoder is composed of $n$ repetitions of a multi-head self-attention layer and a feed-forward network (FFN). Decoder is made up of $n$ stacks of a masked multi-head self-attention layer, a cross-attention layer with the output of encoder, and a FFN. Normalization and residual link (AddNorm) are applied for each layer. To make the input token order-variant, Transformer adds a sinusoidal positional encoding to the input of both encoder and decoder. 

\subsection{3D Object Detection}
\label{sec:method:subsec:detection}
For an input point cloud of size $N\times \left(3 + K\right)$ including a $3$-dim coordinate and extra $K$-dim features such as colors, normals, and multi-view features for each point, we first apply an object detector to generate $M$ object candidates which are input tokens to later components. To make fair comparisons with existing work, we deploy the same seminal detector VoteNet~\cite{votenet} with PointNet++~\cite{pointnet++} as feature aggregation backbone to produce initial object features $\mathcal{X}\in \mathbb{R}^{M\times C}$. We also keep the vote cluster center coordinates $\mathcal{P}\in \mathbb{R}^{M\times 3}$ from its proposal module as global location information for later positional encoding. 

\subsection{Token-to-Token Spatial Relationship Learning}
\label{sec:method:subsec:encoder}
To generate spatiality-enhanced captions, we carefully conduct a relative spatiality modeling from which spatial relations among tokens can be formulated and learnt through a token-to-token (T2T) spatial relationship learning objective. 
% -------------- figure:teaser ---------------
\begin{figure}[!t]
\centering     %%% not \center
\includegraphics[width=\linewidth]{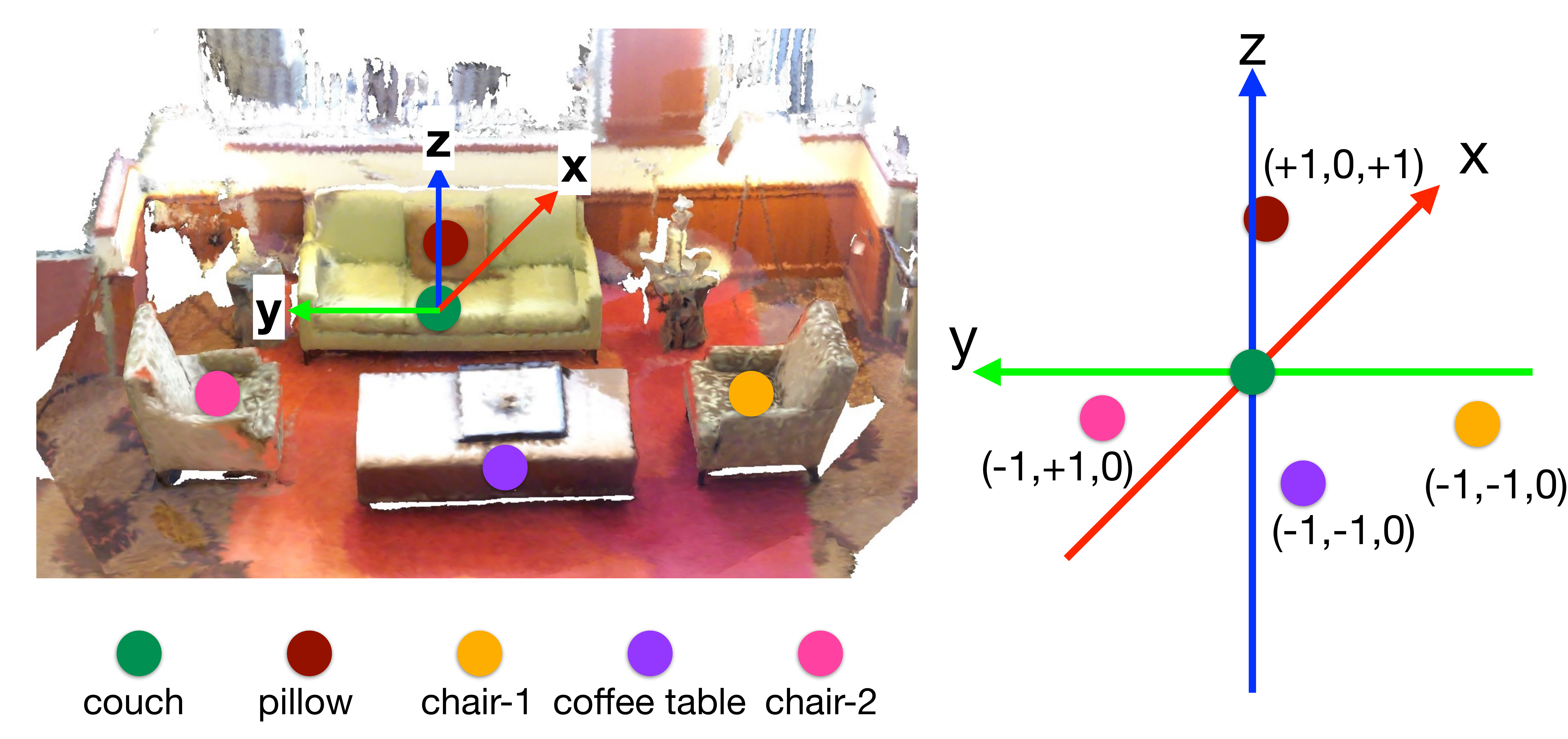}
\caption{An example of our 3D spatiality modeling of main-axis spatial relations. With respect to the couch, chair-2 is in the negative half x-axis, positive half y-axis, and on the same floor along z-axis, hence its spatial relation to couch is represented as (-1, +1, 0). As spatial relation is relative, the relation of couch to chair-2 is expressed reversely as (+1, -1, 0).}
\label{fig:teaser}
\end{figure}
% -------------- figure ---------------
\subsubsection{Relative Spatiality Modeling}
% relative spatial information
We first introduce how we model the relative spatiality in 3D scenes. We construct a local 3D coordinate system (right-handed) with an object itself as the origin center, and the relation of a surrounding object with respect to the center object can be represented as a ($\lambda_x$, $\lambda_y$, $\lambda_z$)-triplet where each entry $\lambda \in \left\{+1,-1,0\right\}$ indicates which half axis the surrounding object sits along different axes (+1 for positive, -1 for negative, and 0 for same position), as illustrated in Figure~\ref{fig:teaser}. Note we only consider coarse direction such as positive or negative and ignore the exact displacement along axes to decomplexify the modeling. Specifically, according to the intersection of two objects, the definition of a positive relation can vary. Before discussing it, we introduce the notations we use in the following. $\Box_i$ indicates the bounding box of an object $o_i$. And $\Box^{k}_{i:\wedge/\vee/ \mid} $ denotes the parameters of the bounding box along $k$-axis where $k\in \left\{x,y,z\right\}$, i.e., $\wedge$ for minimum value, $\vee$ for maximum value, and $\mid$ for side length. As relations along z-axis involve different heights while those along x-/y- axis are grounded to the same floor level, we discuss the criteria of being positive for them separately.

 % -------- figure: positive condition ---------
\begin{figure}[!t]
\centering     %%% not \center
\subfigure[Absolute.]{\includegraphics[width=0.15\textwidth]{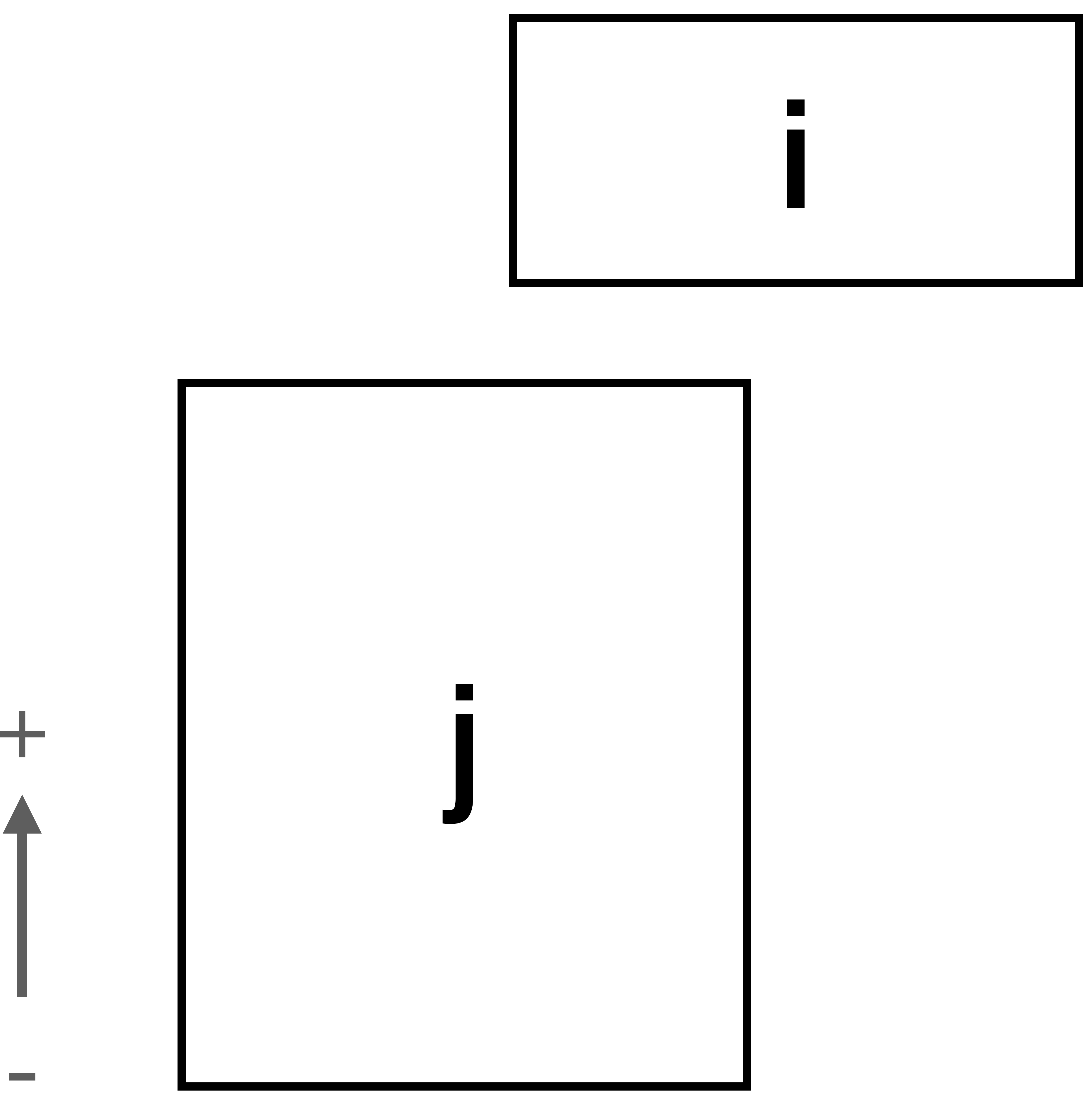}\label{fig:c1}}
\subfigure[Covered.]{\includegraphics[width=0.15\textwidth]{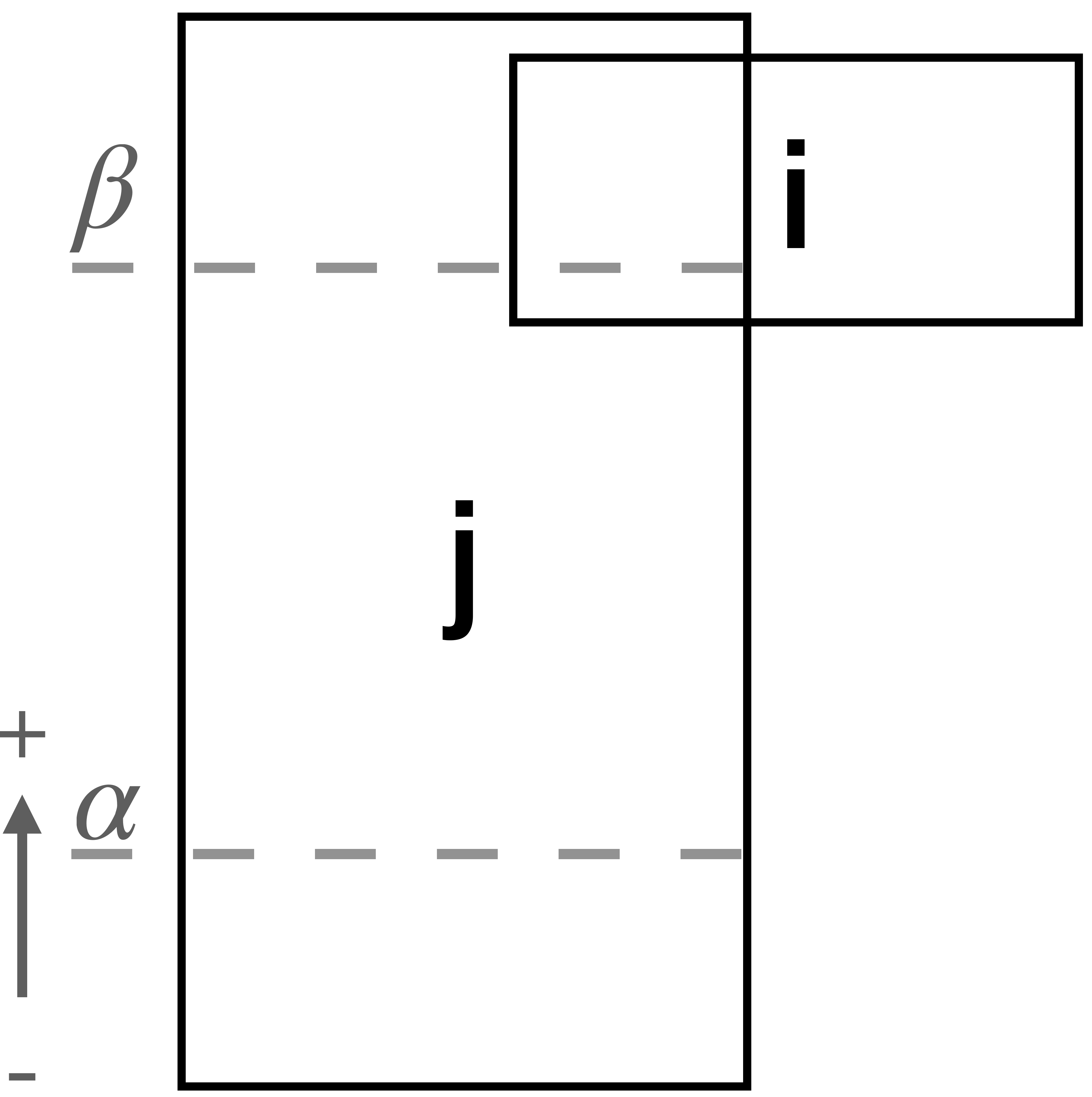}\label{fig:c2}}
\subfigure[Covering.]{\includegraphics[width=0.15\textwidth]{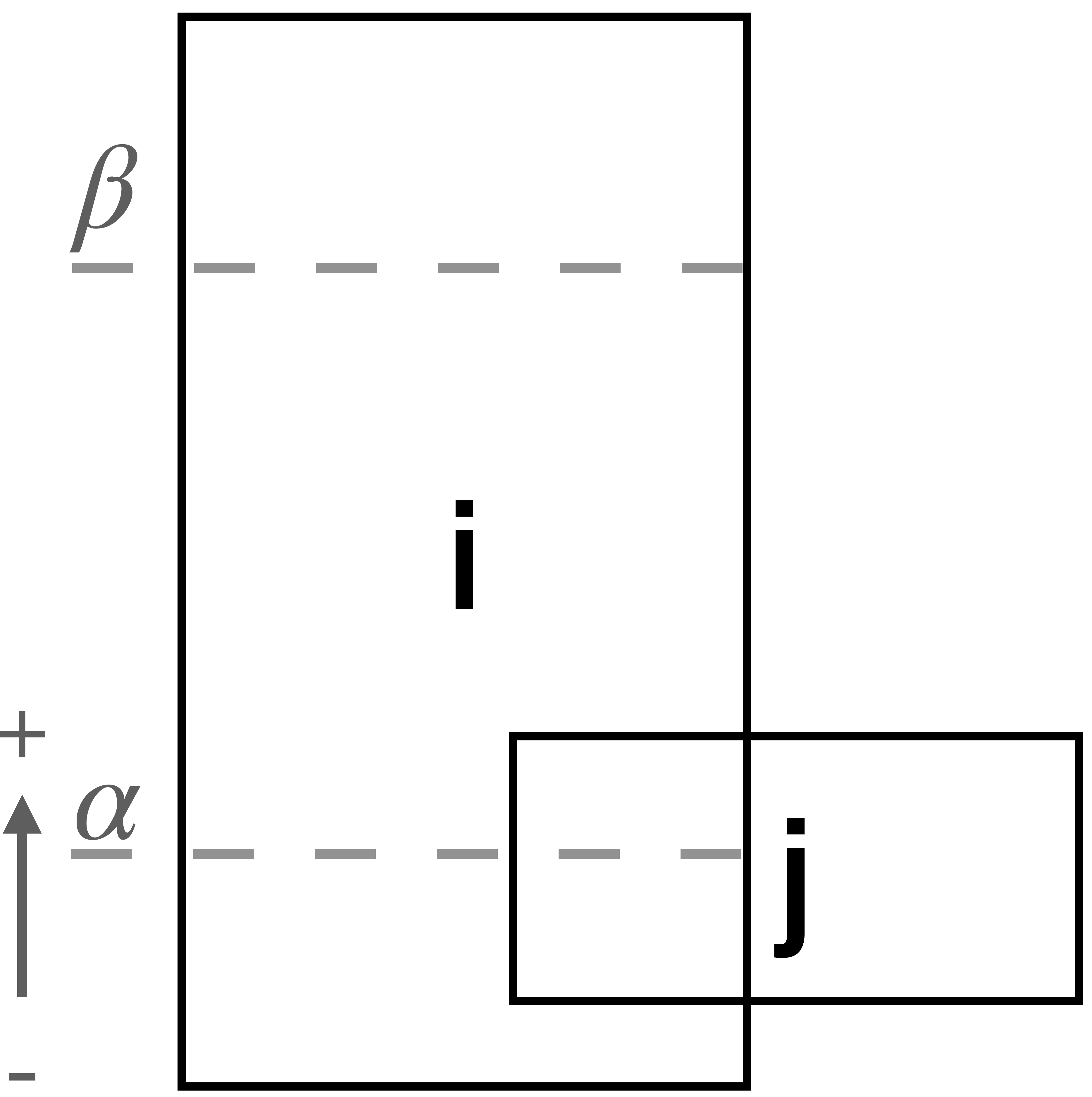}\label{fig:c3}}

\caption{Three cases when an object $o_i$ is to the positive direction of another object $o_j$ along x-/y- axis. Top view. The arrow points to the positive direction. $\alpha$ and $\beta$ are the lower and upper area limits, respectively.}
\label{fig:enc:xy_condition}
\end{figure}
% -------- figure: positive condition ---------
\paragraph{Same floor.} Depending on how two objects overlap with each other, we categorize the criteria of $o_i$ being positive w.r.t. $o_j$ along x-/y- axis into three cases as illustrated in Figure~\ref{fig:enc:xy_condition}. (a) \textbf{Absolute positive}: when the overlapping area does not exceed the side length of $o_j$ and both the bottom and up of $o_i$ are above those of $o_j$: $(\Box^{k}_{i: \wedge} \geq \Box^{k}_{j: \wedge}) \cap (\Box^{k}_{i: \vee} > \Box^{k}_{j: \vee})$ (b) \textbf{Covered positive}: when the overlapping area equals the side length of $o_i$ itself, i.e, $o_i$ is completely covered by $o_j$, and it resides at the upper area of $o_j$: $(\Box^{k}_{i: \wedge} > \Box^{k}_{j: \wedge}+\alpha\times\Box^{k}_{j: \mid}) \cap (\Box^{k}_{j: \wedge}+\beta\times\Box^{k}_{j: \mid} < \Box^{k}_{i: \vee} \leq \Box^{k}_{j: \vee})$ (c) \textbf{Covering positive}: this is the reverse situation of condition (b) when the overlapping area equals the side length of $o_j$ instead, and $o_j$ lags at $o_i$'s lower area: $(\Box^{k}_{i: \wedge} < \Box^{k}_{j: \wedge} < \Box^{k}_{i: \wedge}+\alpha\times\Box^{k}_{i: \mid}) \cap (\Box^{k}_{j: \vee} < \Box^{k}_{i: \wedge}+\beta\times\Box^{k}_{i: \mid})$. We additionally define two objects are at the same position when the bottom and up of one object are within a certain tolerance $\epsilon$ from those of the other one: $(\mid \Box^{k}_{i: \vee} - \Box^{k}_{j: \vee}\mid  \leq \epsilon) \cap (\mid \Box^{k}_{i: \wedge} - \Box^{k}_{j: \wedge}\mid  \leq \epsilon)$. The lower and upper area limits $\alpha$, $\beta$, and tolerance $\epsilon$ are empirically set as 0.3, 0.7, and 0.1$\times \Box^{k}_{j: \mid}$, respectively.

\paragraph{Various heights.} We define $o_i$ is at positive direction to $o_j$ when $o_i$ is over the lower area of $o_j$: $\Box^{z}_{i: \wedge} \geq \Box^{z}_{j: \wedge} + \alpha\times\Box^{z}_{j: \mid}$.

\subsubsection{Positional Encoding}
Before spatial relation learning, as shown in Figure~\ref{fig:overview}, we apply a learnable function $f(\cdot)$ to each vote cluster center $p\in \mathcal{P}$ to incorporate the global location information into each token. Specifically, $f(\cdot)$ is defined as:
\begin{align}
    f(p) = (\sigma(BN(pW_1)))W_2,
\end{align}
where $W_1\in \mathbb{R}^{3\times C}$ and $W_2\in \mathbb{R}^{C\times C}$ are two linear transformations to project 3-dim geometric features into the same high dimensional space as general features $\mathcal{X}$. We use ReLU as the activation function $\sigma$ and a Batch Normalization (BN) layer to adjust the feature distribution. The input tokens $T=\left\{t_1, t_2, ..., t_M\right\}\in\mathbb{R}^{M\times C}$ for our encoder are then created by adding the newly projected $C$-dim global geometry into $\mathcal{X}$.

\subsubsection{Spatial Relation Learning}
\begin{figure}[!t]
\centering     %%% not \center
\includegraphics[width=\linewidth]{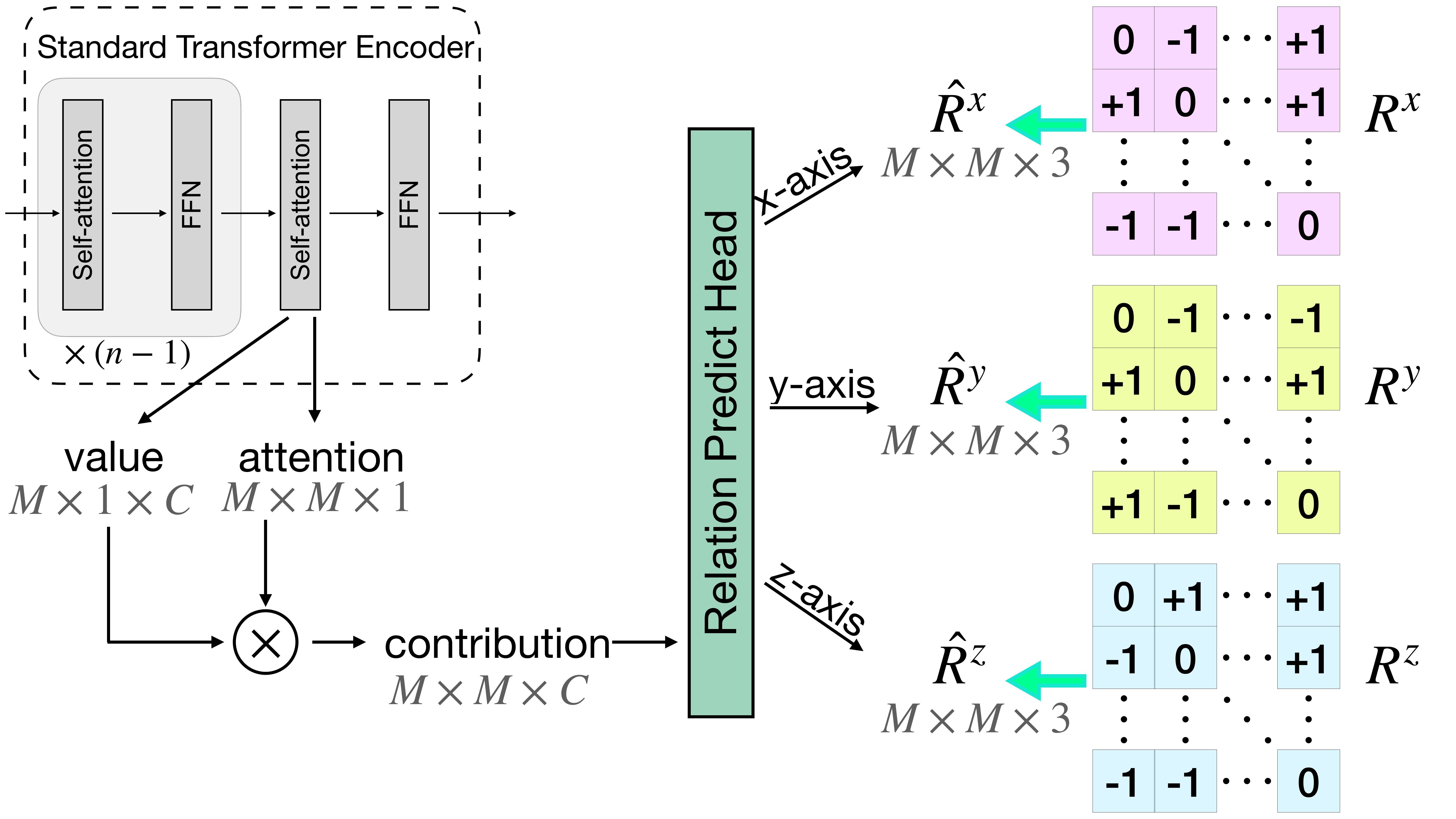}
\caption{Detailed encoder architecture. AddNorm is omitted for brevity.}
\label{fig:encoder}
\end{figure}

When learning a token-to-token spatial relationship, we aim to capture the corresponding object-to-object relation. For a target token $t_i$ and its neighboring token $t_j$, we select their ground truth objects $o_i$ and $o_j$ as the ones with the nearest centers to their predicted centers. We can generate three main-axis spatial relation label maps, $\mathcal{R}^{x}$, $\mathcal{R}^{y}$, and $\mathcal{R}^{z}$, for the $M$ tokens based on their ground truth objects' relations, as per the criteria defined above. Label entries $\left\{r_{i,j}^{x}, r_{i,j}^{y}, r_{i,j}^{z}\right\}\in \left\{+1,-1,0\right\}$ define how the object $o_i$ represented by token $t_i$ is in positive/negative/same direction along x-, y-, and z-axis to another object $o_j$ represented by $t_j$, respectively. In a standard transformer, the encoder is composed of $n$ repetitions of a multi-head self-attention (MSA) layer and a feed-forward network (FFN). A normalization and residual link (AddNorm) is applied for each layer. The updated token $t_i'$ after attention mechanism is defined as the summation of $\omega_{i,j}t_j$ for $j=0,1,...,M$ where $\omega_{i,j}$ represents the attention coefficient of $t_i$ to $t_j$. In other words, the updated token is comprised of different contribution of its neighboring tokens. To encode such contribution with relative 3D spatiality information, we apply our relation prediction head (RPH) to each contribution $\omega_{i,j}t_j$. As illustrated in Figure~\ref{fig:encoder}, the relation prediction happens at the last encoder block. We use a standard three-layer MLP with two ReLU activated $C$-dim hidden layers and a linear output layer. The output of the RPH is a 9-dim vector where each three represents the predicted relation along a main axis. The T2T spatial relation learning is hence guided by our relation loss as:
\begin{align}
    L_{relation} = \sum_{k\in\left\{x,y,z\right\}} L_{CE}(\hat{\mathcal{R}}^{k}, \mathcal{R}^{k}),
\label{formula}
\end{align}
where $L_{CE}$ denotes three-class cross-entropy loss.

% -------------- figure:encoder ---------------
% -------------- figure ---------------

\subsection{Object-centric Decoder}
\label{sec:method:subsec:decoder}
% -------------- figure:decoder ---------------
\begin{figure}[!t]
\centering     %%% not \center
\includegraphics[width=\linewidth]{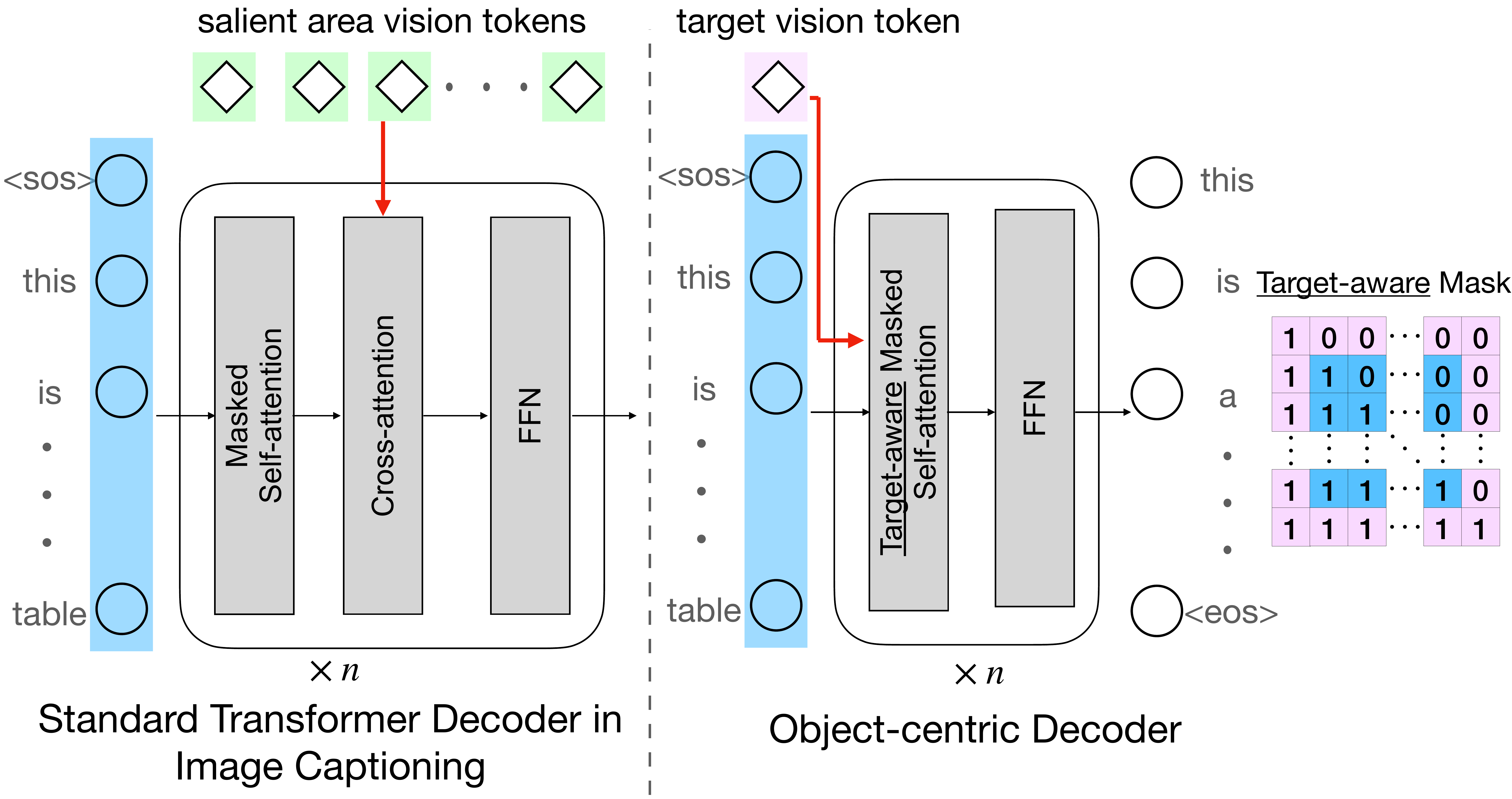}
\caption{Detailed decoder architecture. Positional encoding and AddNorm for decoder are omitted for brevity.}
\label{fig:decoder}
\end{figure}
% -------------- figure ---------------

% ---- sota table begins------
\begin{table*}[!t]
\centering
\resizebox{\textwidth}{!}{%
\begin{tabular}{l|l|l|rrrr|r}
\toprule
Datasets  & Methods & Input & C@0.5IoU & B-4@0.5IoU & M@0.5IoU & R@0.5IoU & mAP@0.5IoU\\
\midrule
ScanRefer & Scan2Cap& xyz+normal+mv & 39.08 &	23.32 &	21.97 &	44.78 &	32.21\\
            & Ours-base & xyz & 40.19 &	24.71 &	22.01 &	45.49 &	32.32\\
            & Ours & xyz &	42.53 &	25.02 &	22.22 &	45.65 &	34.44\\
            &Ours & xyz+normal+rgb & 42.76 & \textbf{25.38} & \textbf{22.84} &	\textbf{45.66} & 35.55\\
            &Ours & xyz+normal+mv & \textbf{44.02} & 25.26 & 22.33 & 45.36 &	\textbf{36.64}\\
\midrule
Nr3D/ReferIt3D & Scan2Cap& xyz+normal+mv &	24.10 &	15.01 &	21.01 &	47.95 &	32.21\\
                   & Ours-base& xyz & 31.06 &	17.94 &	22.03 &	49.63 &	30.65\\
                   & Ours& xyz & 31.43 &	18.98 &	22.24 &	49.79 &	33.17\\
                   & Ours& xyz+normal+rgb & 33.24 & 19.46 & \textbf{22.61} & 50.41 & 33.23\\
                   & Ours& xyz+normal+mv & \textbf{33.71} & \textbf{19.92} & 22.61 & \textbf{50.50} & \textbf{38.11}\\

\bottomrule
\end{tabular}}
\caption{Quantitative comparison with SOTA methods on ScanRefer and Nr3D/ReferIt3D. Ours-base is the baseline variant of standard Transformer adapted for 3D dense captioning, where we use standard encoder with sinusoidal positional encoding and late-guide decoder. The input denotes various combinations of different information: \textit{xyz} refers to points' coordinates. \textit{normal} means the normal vector of each point. \textit{rgb} uses color information and \textit{mv} stands for pretrained 2D multi-view features.}
\label{tab:sota}
\end{table*}
In image captioning Transformers, the decoder consisting of $n$ stacks of a masked MSA layer, a cross-attention layer with the output of encoder, and a FFN, attends all salient area vision tokens to conclude one sentence describing the whole image. On the other hand, in dense captioning, the target is each object. Therefore, we propose an object-centric decoder with target-aware masked self-attention layer to update each word token by attending both its previous words and the target vision token as depicted in Figure~\ref{fig:decoder}. Compared to the standard decoder, our design can fulfill the dense captioning task but in a more concise and efficient manner. More specifically, it would stack a target vision token mask (in pink) on the basis of the existing word token mask (in blue) and feed the target vision token as well as word tokens together into the self-attention layer. Considering ours as an early-guide way to condition the decoder on the target object, we also implemented a late-guide variant and have it ablated in Table~\ref{tab:abl1}.
% We also tried methods such as adding/concating each input word token with the target token but resulted in strong overfitting problem.

\subsection{Learning Objective}
\label{sec:method:subsec:loss}
We define our final loss as $L = \delta*L_{det} + L_{des} + \zeta*L_{relation}$, where $L_{relation}$ is our proposed T2T spatial relation learning objective defined in Equation~\ref{formula}. As for the object detection loss $L_{det}$ and the description loss $L_{des}$, we follow Scan2Cap, and more details can be found in ~\cite{scan2cap}. $\delta$ and $\zeta$ are set as 10 and 0.1, respectively, to maintain similar magnitude of different losses.

% -------------- figure:qualitative ---------------
\begin{figure*}[!t]
\centering     %%% not \center
\includegraphics[width=\textwidth]{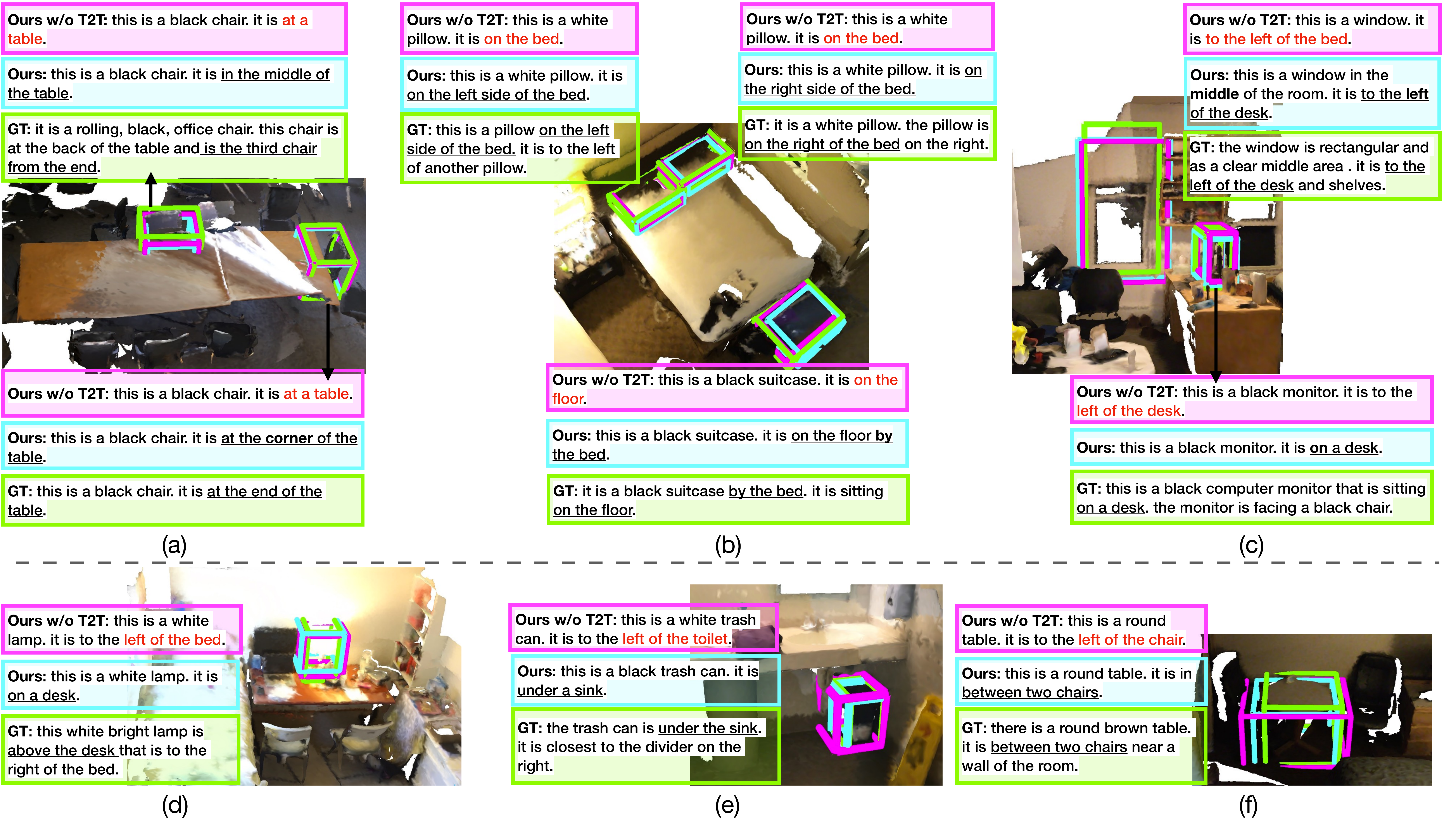}
\caption{Visualization for our method with and without token-to-token (T2T) spatial relation guidance. Caption boxes share the same color with detection bounding boxes for ground truth (green), ours w/ T2T (blue), and ours w/o T2T (pink). Imprecise parts of sentences produced by ours w/o T2T guidance are marked in red, and correctly expressed spatial relations predicted by T2T-guided method are underscored.}
\label{fig:qualitivie}
\end{figure*}
% -------------- figure ---------------

\section{Experiments}

\subsection{Datasets, Metrics, and Implementation Details}
\paragraph{Datasets.}
We evaluate our proposed method on ScanRefer~\cite{scanrefer} and Nr3D from ReferIt3D~\cite{referit3d}, both of which provide free-form human descriptions for objects in ScanNet~\cite{scannet}. Same as Scan2Cap~\cite{scan2cap}, for ScanRefer/Nr3D, we train on 36,665/32,919 captions for 7,875/4,664 objects from 562/511 scenes and evaluate on 9,508/8,584 descriptions for 2,068/1,214 objects from 141/130 scenes.

\paragraph{Metrics.}
We benchmark the performances on both detection and captioning perspectives. For detection, we use the mean average precision thresholded by 0.5 IoU score (mAP$@$0.5). For captioning, we employ $m@0.5$IoU where only the prediction whose IoU is larger than 0.5 will be considered~\cite{scan2cap}. The captioning metric $m$ can be the one especially designed for image captioning such as CIDEr (C)~\cite{cider}, or those focusing more on machine translation or on text summarizing such as BLEU-4 (B-4)~\cite{bleu}, METEOR (M)~\cite{meteor}, and ROUGE (R)~\cite{rouge}.

\paragraph{Implementation Details.}
To make fair comparisons, we use the same training and testing protocols as Scan2Cap. Following Scan2Cap, we set the input number of points $N$ as 40,000 and the number of object proposal $M$ as 256. The output feature dimension $C$ from object detector is 128. For Transformer, we set the number of encoder and decoder blocks $n$ as 6 and the number of heads in multi-head attentions as 8. The dimensionality of input and output of each layer is 128 except that for the inner-layer of feed-forward networks as 2048. We implement AddNorm as pre-LN where outputs are first normalized and then added with inputs. Keeping the length of descriptions within 30 tokens and marking words not appearing in GloVE~\cite{glove} as unknown, we learn the word embedding from scratch encoded with sinusoidal positional encoding. We implement\footnote{\url{https://github.com/heng-hw/SpaCap3D}} our proposed model in PyTorch~\cite{pytorch} and train end-to-end with ADAM~\cite{adam} in a learning rate of $1\times10^{-3}$. The detection backbone is fine-tuned together with the end-to-end training from the pretrained VoteNet~\cite{votenet} model provided by Scan2Cap. To avoid overfitting, we apply the same weight decay factor $1\times10^{-5}$ and the same data augmentation as Scan2Cap. We firstly randomly flip the input point cloud along the YZ- and XZ-plane and then rotate along x-, y-, and z-axis by a random angle within $\left[-5^{\circ},5^{\circ}\right]$. We finally translate the point cloud along all axes by a random distance within $\left[-0.5,0.5\right]$ meters. All experiments were trained on a single GeForce RTX 2080Ti GPU with a batch size of 8 samples for 50 epochs, while the model is checked and saved when it reaches the best CIDEr@0.5IoU on val split every 2000 iterations. Training with our proposed framework takes around 33 and 29 hours for ScanRefer and Nr3D/ReferIt3D, respectively. During inference, we use non-maximum suppression to remove overlapping proposals and only keep those proposals whose IoUs with ground truth bounding boxes are larger than 0.5. With a batch size of 8, the inference time including evaluation with the four captioning metrics, for ScanRefer and Nr3D/ReferIt3D, is 83 and 75 seconds, respectively. We implement the attention visualization shown in Figure~\ref{fig:newteaser} and the supplementary based on ~\cite{bertviz}.
% More details can be found in our supplementary document. 

\subsection{Quantitative Comparison}
Table~\ref{tab:sota} presents the quantitative comparison with SOTA methods, showing our proposed SpaCap3D with xyz-input outperforms not only the baseline Scan2Cap but also the standard Transformer (ours-base) in all metrics. It is worth noting that SpaCap3D manages to exceed Scan2Cap, even when SpaCap3D only takes simple coordinates as inputs whereas Scan2Cap uses much richer pretrained multi-view features. To be comparable with Scan2Cap's input setting, we also provide a variant of our proposed method with color and normal features added and it achieves better results as expected. The inclusion of multi-view features can further boost the performance and we point out that it requires 6 more hours to train compared to its simpler counterpart (i.e., xyz+normal+rgb).

The average forward pass time per training batch is around 0.5s and 0.2s, and the average per-batch inference time is around 11.7s and 2.3s, respectively for Scan2Cap and our SpaCap3D (taking the same xyz-input), which demonstrates our efficiency.

\subsection{Qualitative Analysis}

To visualize the importance of relative spatiality, we display some detect-and-describe results of the proposed method with and without relative 3D spatiality learning in Figure~\ref{fig:qualitivie}. If the T2T guidance is discarded, the generated descriptions could lack unique spatial relations and tend to be general as shown in Figure~\ref{fig:qualitivie} (a) and (b). While our spatiality-guided Transformer distinguishes two chairs at a table (\textit{``middle"} and \textit{``corner"}) and two pillows on the bed (\textit{``left"} and \textit{``right"}) by their spatiality, the method without such guidance could collapse into generic expressions lacking specific spatial relations. Also in Figure~\ref{fig:qualitivie} (b), our proposed method with T2T is capable of describing more relations for the suitcase compared to the one without T2T guidance (\textit{``on the floor by the bed"} vs. \textit{``on the floor"}). Figure~\ref{fig:qualitivie} (c), (d), and (e) show cases when T2T guidance boosts correct spatial relation prediction. We also emphasize Figure~\ref{fig:qualitivie} (f) where a table is in between two chairs. Instead of just describing the relation between the table and one chair, the T2T-guided method considers the existence of both chairs and generates more thoughtful expression.  More results are displayed in Supplementary Figure~\ref{fig:more_qualitivie}.

% --- Ablation study table 1 begins---
\begin{table*}[!t]
\centering
\begin{tabular}{l|llll|r|r}
\toprule
\multirow{2}{*}{Model} & \multicolumn{2}{c|}{Decoder} & \multirow{2}{*}{Encoder}  & \multirow{2}{*}{T2T}  & 
\multirowcell{2}{C@0.5IoU\\ \textit{(captioning)}} & \multirowcell{2}{mAP@0.5IoU\\ \textit{\ \quad(detection)}}\\
\cmidrule{2-3}
      & late-guide & \multicolumn{1}{c|}{early-guide}  &  &  &  &  \\
     
\midrule
A   & \checkmark &	& & & 37.80 &	30.97 \\
B   &  & \checkmark	& & & 41.36 &	32.14 \\
C   &  & \checkmark	& \checkmark &  & 41.14 & 33.45 \\
D  &  & \checkmark &  \checkmark & \checkmark	 & \textbf{42.53} &	\textbf{34.44}\\

\bottomrule
\end{tabular}
\caption{Ablation study on different components of our proposed method. T2T denotes the token-to-token spatial relation learning objective.}
\label{tab:abl1}
\end{table*}
% --- Ablation study table 1 end---

\subsection{Ablation Study}

\subsubsection{Component Analysis}
We investigate components of our proposed architecture, the late-guide and early-guide decoder, attention-based encoder with vote center-based positional encoding, and token-to-token spatial relation learning (T2T), in Table~\ref{tab:abl1}. Model A and B adopt the decoder alone and the outcome that Model B achieves 3.56\% and 1.17\% improvement over Model A on captioning and detection respectively demonstrates the superiority of our proposed early-guide decoder. Based on Model B, Model C uses an attention-based encoder to learn the long-range dependency among object proposals, which leads to a detection performance increase by 1.31\%. With the guidance of our T2T spatial relation learning objective, the encoder functions better as can be seen in the results from Model D which performs the best in both captioning and detection.

% --- Ablation study table 2 begins---
\begin{table}[!tbh]
\centering
\begin{tabular}{l|l|r}
\toprule
\multicolumn{2}{c|}{Positional encoding}  & C@0.5IoU \\
\midrule
non-learnable   & none &	39.41 \\
            & sinusoidal &	39.44\\
\midrule
learnable    & random &	42.29\\
            & box center &	42.49 \\
            & box center* &	40.04\\
            & vote center &	\textbf{42.53}\\
\bottomrule
\end{tabular}
\caption{Ablation study on choices of encoder's positional encoding. box center$*$ indicates concatenation of box center and size.}
\label{tab:abl2}
\end{table}
% --- Ablation study table 2 end---
\subsubsection{Positional Encoding Analysis}
To verify the choice of learnable vote center-based positional encoding for encoder, we elaborate on different ways in Table~\ref{tab:abl2}. The non-learnable sinusoidal method in standard Transformer has slightly better effect over the one without any positional encoding, showing the necessity of such encoding in our task. For learnable encoding, we compare with random one used in 2D object detection Transformer~\cite{detr} and box center-based one adopted in 3D object detection Transformer~\cite{groupfree}. More details of these learnable positional encoding methods can be found in the supplementary. We find that random learnable positional encoding can boost the performance compared to non-learnable ones and the incorporation of object position information can further advance the performance. Among all learnable encoding ways, our vote center-based one achieves the best results.
% We illustrate these three different learnable positional encoding methods in Figure~\ref{fig:abl2_illustrate}. Random approach is based on nothing and fixed during inference. Box center and vote center are both based on the position parameter of object proposals and therefore dynamic at inference. As shown in Figure~\ref{fig:abl2_illustrate}, box center indicates the center of an object token's predicted bounding box (marked in green) while the vote center refers to the center of vote cluster (marked in red) from which a token is proposed via aggregation. We find that random learnable positional encoding can boost the performance compared to non-learnable ones and the incorporation of object position information can further advance the performance. Among all learnable encoding ways, our vote center-based one achieves the best.

\section{Conclusion}
In this work, we propose a new state-of-the-art framework dubbed as SpaCap3D for the newly emerging 3D dense captioning task. We propose to formulate object relations with relative 3D spatiality modeling, based on which we build a transformer-based architecture where a spatiality-guided encoder learns how objects interact with their surrounding environment in 3D spatiality via a token-to-token spatial relation learning guidance, and a shared object-centric decoder is conditioned on each spatiality-enhanced token to individually generate precise and unambiguous object-level captions. Experiments on two benchmark datasets show that our integrated framework outperforms the baseline method by a great deal in both accuracy and efficiency.

% \section*{Acknowledgments}
% TBD.
% \appendix

% \section{\LaTeX{} and Word Style Files}\label{stylefiles}
% TBD.

\bibliographystyle{named}
\bibliography{ijcai22}

\begin{thebibliography}{}

\bibitem[\protect\citeauthoryear{Achlioptas \bgroup \em et al.\egroup
  }{2020}]{referit3d}
Panos Achlioptas, Ahmed Abdelreheem, Fei Xia, Mohamed Elhoseiny, and Leonidas
  Guibas.
\newblock Referit3{D}: Neural listeners for fine-grained 3{D} object
  identification in real-world scenes.
\newblock In {\em ECCV}, pages 422--440. Springer, 2020.

\bibitem[\protect\citeauthoryear{Avetisyan \bgroup \em et al.\egroup
  }{2019}]{scan2cad}
Armen Avetisyan, Manuel Dahnert, Angela Dai, Manolis Savva, Angel~X Chang, and
  Matthias Nie{\ss}ner.
\newblock Scan2{CAD}: Learning {CAD} model alignment in {RGB-D} scans.
\newblock In {\em CVPR}, pages 2614--2623, 2019.

\bibitem[\protect\citeauthoryear{Banerjee and Lavie}{2005}]{meteor}
Satanjeev Banerjee and Alon Lavie.
\newblock {METEOR}: An automatic metric for {MT} evaluation with improved
  correlation with human judgments.
\newblock In {\em {ACL} {W}orkshop: {I}ntrinsic and {E}xtrinsic {E}valuation
  {M}easures for {M}achine {T}ranslation and/or {S}ummarization}, pages 65--72,
  2005.

\bibitem[\protect\citeauthoryear{Carion \bgroup \em et al.\egroup
  }{2020}]{detr}
Nicolas Carion, Francisco Massa, Gabriel Synnaeve, Nicolas Usunier, Alexander
  Kirillov, and Sergey Zagoruyko.
\newblock End-to-end object detection with transformers.
\newblock In {\em ECCV}, pages 213--229. Springer, 2020.

\bibitem[\protect\citeauthoryear{Chen \bgroup \em et al.\egroup
  }{2020}]{scanrefer}
Dave~Zhenyu Chen, Angel~X Chang, and Matthias Nie{\ss}ner.
\newblock Scan{R}efer: 3{D} object localization in {RGB-D} scans using natural
  language.
\newblock In {\em ECCV}, pages 202--221. Springer, 2020.

\bibitem[\protect\citeauthoryear{Chen \bgroup \em et al.\egroup
  }{2021}]{scan2cap}
Zhenyu Chen, Ali Gholami, Matthias Nie{\ss}ner, and Angel~X Chang.
\newblock Scan2{C}ap: Context-aware dense captioning in {RGB-D} scans.
\newblock In {\em CVPR}, pages 3193--3203, 2021.

\bibitem[\protect\citeauthoryear{Cornia \bgroup \em et al.\egroup }{2020}]{m2}
Marcella Cornia, Matteo Stefanini, Lorenzo Baraldi, and Rita Cucchiara.
\newblock Meshed-memory transformer for image captioning.
\newblock In {\em CVPR}, pages 10578--10587, 2020.

\bibitem[\protect\citeauthoryear{Dai \bgroup \em et al.\egroup
  }{2017}]{scannet}
Angela Dai, Angel~X Chang, Manolis Savva, Maciej Halber, Thomas Funkhouser, and
  Matthias Nie{\ss}ner.
\newblock Scan{N}et: Richly-annotated 3{D} reconstructions of indoor scenes.
\newblock In {\em CVPR}, pages 5828--5839, 2017.

\bibitem[\protect\citeauthoryear{Herdade \bgroup \em et al.\egroup
  }{2019}]{relationtransformer}
Simao Herdade, Armin Kappeler, Kofi Boakye, and Joao Soares.
\newblock Image captioning: transforming objects into words.
\newblock In {\em NeurIPS}, pages 11137--11147, 2019.

\bibitem[\protect\citeauthoryear{Ji \bgroup \em et al.\egroup
  }{2021}]{ji2021improving}
Jiayi Ji, Yunpeng Luo, Xiaoshuai Sun, Fuhai Chen, Gen Luo, Yongjian Wu, Yue
  Gao, and Rongrong Ji.
\newblock Improving image captioning by leveraging intra-and inter-layer global
  representation in transformer network.
\newblock In {\em AAAI}, volume~35, pages 1655--1663, 2021.

\bibitem[\protect\citeauthoryear{Johnson \bgroup \em et al.\egroup
  }{2016}]{densecap}
Justin Johnson, Andrej Karpathy, and Li~Fei-Fei.
\newblock Dense{C}ap: Fully convolutional localization networks for dense
  captioning.
\newblock In {\em CVPR}, pages 4565--4574, 2016.

\bibitem[\protect\citeauthoryear{Kim \bgroup \em et al.\egroup }{2019}]{2ddc4}
Dong-Jin Kim, Jinsoo Choi, Tae-Hyun Oh, and In~So Kweon.
\newblock Dense relational captioning: Triple-stream networks for
  relationship-based captioning.
\newblock In {\em CVPR}, pages 6271--6280, 2019.

\bibitem[\protect\citeauthoryear{Kingma and Ba}{2015}]{adam}
Diederik~P Kingma and Jimmy Ba.
\newblock Adam: A method for stochastic optimization.
\newblock In {\em ICLR}, 2015.

\bibitem[\protect\citeauthoryear{Landau and Jackendoff}{1993}]{spatial1}
Barbara Landau and Ray Jackendoff.
\newblock ``{W}hat" and ``where" in spatial language and spatial cognition.
\newblock {\em Behavioral and Brain Sciences}, 16(2):255--265, 1993.

\bibitem[\protect\citeauthoryear{Li \bgroup \em et al.\egroup
  }{2019a}]{entangle}
Guang Li, Linchao Zhu, Ping Liu, and Yi~Yang.
\newblock Entangled transformer for image captioning.
\newblock In {\em ICCV}, pages 8928--8937, 2019.

\bibitem[\protect\citeauthoryear{Li \bgroup \em et al.\egroup }{2019b}]{2ddc3}
Xiangyang Li, Shuqiang Jiang, and Jungong Han.
\newblock Learning object context for dense captioning.
\newblock In {\em AAAI}, volume~33, pages 8650--8657, 2019.

\bibitem[\protect\citeauthoryear{Lin}{2004}]{rouge}
Chin-Yew Lin.
\newblock Rouge: A package for automatic evaluation of summaries.
\newblock In {\em {ACL} {W}orkshop: {T}ext {S}ummarization {B}ranches {O}ut},
  pages 74--81, 2004.

\bibitem[\protect\citeauthoryear{Liu \bgroup \em et al.\egroup
  }{2021}]{groupfree}
Ze~Liu, Zheng Zhang, Yue Cao, Han Hu, and Xin Tong.
\newblock Group-free 3{D} object detection via transformers.
\newblock In {\em ICCV}, pages 2949--2958, 2021.

\bibitem[\protect\citeauthoryear{Luo \bgroup \em et al.\egroup
  }{2021}]{luo2021dual}
Yunpeng Luo, Jiayi Ji, Xiaoshuai Sun, Liujuan Cao, Yongjian Wu, Feiyue Huang,
  Chia-Wen Lin, and Rongrong Ji.
\newblock Dual-level collaborative transformer for image captioning.
\newblock In {\em AAAI}, volume~35, pages 2286--2293, 2021.

\bibitem[\protect\citeauthoryear{Papineni \bgroup \em et al.\egroup
  }{2002}]{bleu}
Kishore Papineni, Salim Roukos, Todd Ward, and Wei-Jing Zhu.
\newblock Bleu: a method for automatic evaluation of machine translation.
\newblock In {\em ACL}, pages 311--318, 2002.

\bibitem[\protect\citeauthoryear{Paszke \bgroup \em et al.\egroup
  }{2019}]{pytorch}
Adam Paszke, Sam Gross, Francisco Massa, Adam Lerer, James Bradbury, Gregory
  Chanan, Trevor Killeen, Zeming Lin, Natalia Gimelshein, Luca Antiga, et~al.
\newblock Pytorch: An imperative style, high-performance deep learning library.
\newblock In {\em NeurIPS}, pages 8026--8037, 2019.

\bibitem[\protect\citeauthoryear{Pennington \bgroup \em et al.\egroup
  }{2014}]{glove}
Jeffrey Pennington, Richard Socher, and Christopher~D Manning.
\newblock Glove: Global vectors for word representation.
\newblock In {\em EMNLP}, pages 1532--1543, 2014.

\bibitem[\protect\citeauthoryear{Qi \bgroup \em et al.\egroup
  }{2017}]{pointnet++}
Charles~R Qi, Li~Yi, Hao Su, and Leonidas~J Guibas.
\newblock Pointnet++: Deep hierarchical feature learning on point sets in a
  metric space.
\newblock In {\em NeurIPS}, 2017.

\bibitem[\protect\citeauthoryear{Qi \bgroup \em et al.\egroup
  }{2018}]{qi2018frustum}
Charles~R Qi, Wei Liu, Chenxia Wu, Hao Su, and Leonidas~J Guibas.
\newblock Frustum pointnets for 3{D} object detection from {RGB-D} data.
\newblock In {\em CVPR}, pages 918--927, 2018.

\bibitem[\protect\citeauthoryear{Qi \bgroup \em et al.\egroup }{2019}]{votenet}
Charles~R Qi, Or~Litany, Kaiming He, and Leonidas~J Guibas.
\newblock Deep hough voting for 3{D} object detection in point clouds.
\newblock In {\em ICCV}, pages 9277--9286, 2019.

\bibitem[\protect\citeauthoryear{Skubic \bgroup \em et al.\egroup
  }{2004}]{spatial2}
Marjorie Skubic, Dennis Perzanowski, Samuel Blisard, Alan Schultz, William
  Adams, Magda Bugajska, and Derek Brock.
\newblock Spatial language for human-robot dialogs.
\newblock {\em IEEE Transactions on Systems, Man, and Cybernetics, Part C
  (Applications and Reviews)}, 34(2):154--167, 2004.

\bibitem[\protect\citeauthoryear{Uppal \bgroup \em et al.\egroup
  }{2022}]{multimodal}
Shagun Uppal, Sarthak Bhagat, Devamanyu Hazarika, Navonil Majumder, Soujanya
  Poria, Roger Zimmermann, and Amir Zadeh.
\newblock Multimodal research in vision and language: A review of current and
  emerging trends.
\newblock {\em Information Fusion}, 77:149--171, 2022.

\bibitem[\protect\citeauthoryear{Vaswani \bgroup \em et al.\egroup
  }{2017}]{transformer}
Ashish Vaswani, Noam Shazeer, Niki Parmar, Jakob Uszkoreit, Llion Jones,
  Aidan~N Gomez, {\L}ukasz Kaiser, and Illia Polosukhin.
\newblock Attention is all you need.
\newblock In {\em NeurIPS}, pages 5998--6008, 2017.

\bibitem[\protect\citeauthoryear{Vedantam \bgroup \em et al.\egroup
  }{2015}]{cider}
Ramakrishna Vedantam, C~Lawrence~Zitnick, and Devi Parikh.
\newblock Cider: Consensus-based image description evaluation.
\newblock In {\em CVPR}, pages 4566--4575, 2015.

\bibitem[\protect\citeauthoryear{Vig}{2019}]{bertviz}
Jesse Vig.
\newblock A multiscale visualization of attention in the transformer model.
\newblock In {\em ACL: System Demonstrations}, pages 37--42, 2019.

\bibitem[\protect\citeauthoryear{Wald \bgroup \em et al.\egroup
  }{2020}]{scenegraph}
Johanna Wald, Helisa Dhamo, Nassir Navab, and Federico Tombari.
\newblock Learning 3{D} semantic scene graphs from 3{D} indoor reconstructions.
\newblock In {\em CVPR}, pages 3961--3970, 2020.

\bibitem[\protect\citeauthoryear{Yang \bgroup \em et al.\egroup }{2017}]{2ddc1}
Linjie Yang, Kevin Tang, Jianchao Yang, and Li-Jia Li.
\newblock Dense captioning with joint inference and visual context.
\newblock In {\em CVPR}, pages 2193--2202, 2017.

\bibitem[\protect\citeauthoryear{Yin \bgroup \em et al.\egroup }{2019}]{2ddc2}
Guojun Yin, Lu~Sheng, Bin Liu, Nenghai Yu, Xiaogang Wang, and Jing Shao.
\newblock Context and attribute grounded dense captioning.
\newblock In {\em CVPR}, pages 6241--6250, 2019.

\bibitem[\protect\citeauthoryear{Zhang \bgroup \em et al.\egroup
  }{2021}]{scenegraph2}
Chaoyi Zhang, Jianhui Yu, Yang Song, and Weidong Cai.
\newblock Exploiting edge-oriented reasoning for 3{D} point-based scene graph
  analysis.
\newblock In {\em CVPR}, pages 9705--9715, 2021.

\bibitem[\protect\citeauthoryear{Zhao \bgroup \em et al.\egroup }{2021}]{3dvg}
Lichen Zhao, Daigang Cai, Lu~Sheng, and Dong Xu.
\newblock {3DVG}-{T}ransformer: Relation modeling for visual grounding on point
  clouds.
\newblock In {\em ICCV}, pages 2928--2937, 2021.

\end{thebibliography}

\clearpage
\begin{appendices}

In this supplementary for SpaCap3D, we provide more details of the learnable positional encoding in Section~\ref{sec:pos_enc}. We visualize the attention mechanism used in our SpaCap3D framework in Section~\ref{sec:attn} and provide more qualitative results of our method in Section~\ref{sec:quali}.

\section{Learnable Positional Encoding}
\label{sec:pos_enc}

\begin{figure}[!thb]
\centering     %%% not \center
\includegraphics[width=0.7\linewidth]{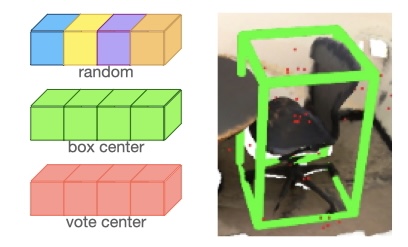}
\caption{Illustration of different ways of positional encoding. \textit{random} refers to the randomly learnable weights, while \textit{box center} and \textit{vote center} use features originated from detected box center (and its size optionally) and vote cluster center, respectively. \textit{vote center} achieves the best results in Table~\ref{tab:abl2} of our main paper.}
\label{fig:abl2_illustrate}
\end{figure}

Figure~\ref{fig:abl2_illustrate} illustrates the three different learnable positional encoding approaches for tokens to the encoder. To generate the $C$-dim positional encoding vector for each token input to the encoder, the random one, as used in 2D detection Transformer~\cite{detr}, randomly learns weight parameters during training and such learnt weights are used as positional encoding for $M$ proposals/tokens during inference, which are fixed for different scene inputs. To make positional encoding object-variant, ~\cite{groupfree} further proposed to generate positional encoding based on predicted box parameters, box center and box size optionally, as shown as the green bounding box in Figure~\ref{fig:abl2_illustrate}. We implement such approach by directly using the predicted bounding box center and size from detection backbone for each proposal. The vote center-based way is similar to the box center-based one but vote centers are the $M$ centers after grouping in proposal module from the detection backbone. Red dots in Figure~\ref{fig:abl2_illustrate} refer to the votes after voting module from which vote centers are generated using farthest point sampling technique.
% -------------- figure:ab2illustration ---------------
% -------------- figure ---------------

\section{Attention Visualization}
\label{sec:attn}
\begin{figure*}[!htb]
\centering     %%% not \center
\subfigure[]{\includegraphics[width=0.6\textwidth]{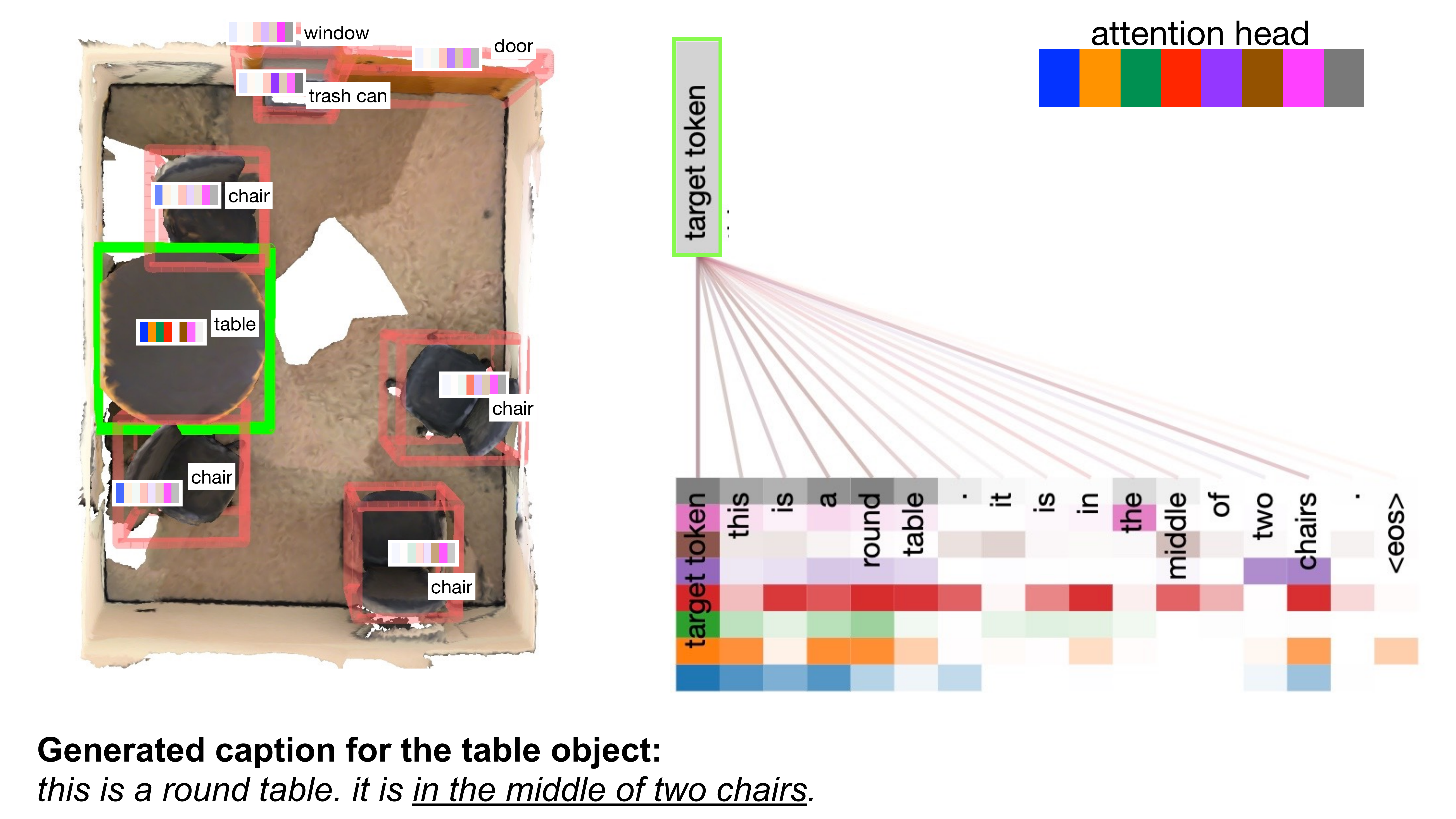}\label{fig:attn_vis1}}
\subfigure[]{\includegraphics[width=0.6\textwidth]{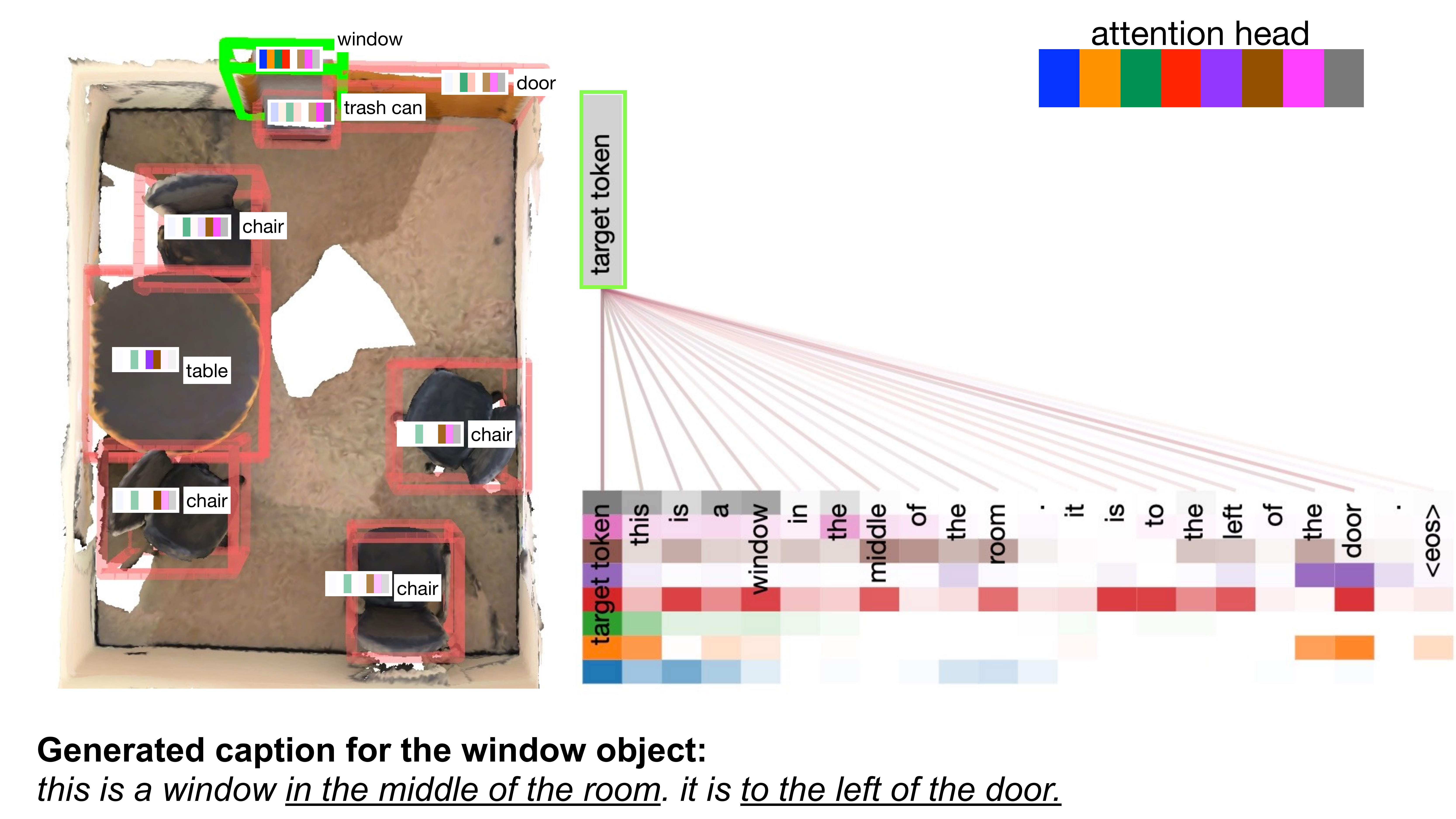}\label{fig:attn_vis2}}
\subfigure[]{\includegraphics[width=0.6\textwidth]{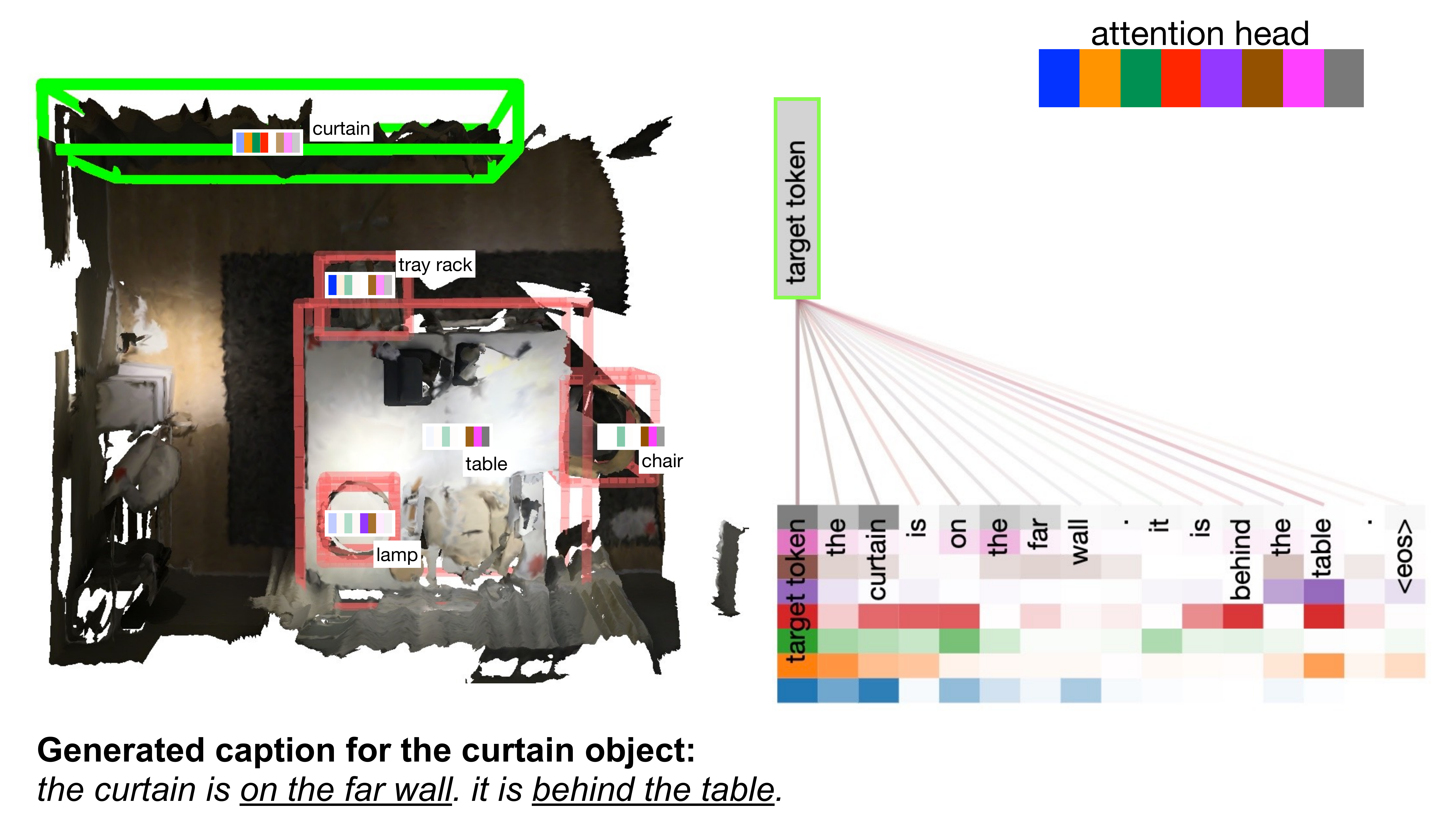}\label{fig:attn_vis3}}

\caption{Examples of our encoder and decoder attention for a target object marked in green in a 3D scene. In each eight-color vector, different colors represent different attention heads. The more transparent the color is, the smaller the attention value is. The colorful vector shown on each object represents the eight-head attention values between the target object marked in green and its surrounding objects in red. The decoder attention between the target vision token and the generated caption words is shown as the eight-color vector underlying each generated word.}
\label{fig:attn_vis}
\end{figure*}

We provide three examples of how the attention works in our proposed method in Figure~\ref{fig:attn_vis}. We follow~\cite{bertviz} to visualize the attentions which are extracted from the last block in both encoder and decoder and values from different heads (eight in total) are marked with different colors. We use opacity to represent the magnitude. The more transparent the color, the smaller the value. In each example, the target object to be described is highlighted in green and the surrounding objects are marked in red. The left figure in each example shows how the surrounding objects contribute to the target object representation learning. As the spatial relations between the target object and its neighbors are different, attentions learnt for different surrounding objects are different. We also present how the spatiality-enhanced target vision token contributes to the generation of each predicted word in the right figure of each example. Taking Figure~\ref{fig:attn_vis1} for instance, we observe that our target vision token contributes differently to the predicted words. It especially emphasizes the words describing the target object itself (\textit{``round table"}), the words expressing the spatial relation (\textit{``the middle of"}), and the words about the neighboring objects (\textit{``two chairs"}), which demonstrates the successful incorporation of relative 3D spatiality in the representation learning phase through our proposed spatiality-guided encoder.

\section{More Qualitative Results}
\label{sec:quali}
\begin{figure*}[!htb]
\centering     %%% not \center
\subfigure[]{\includegraphics[width=0.6\textwidth]{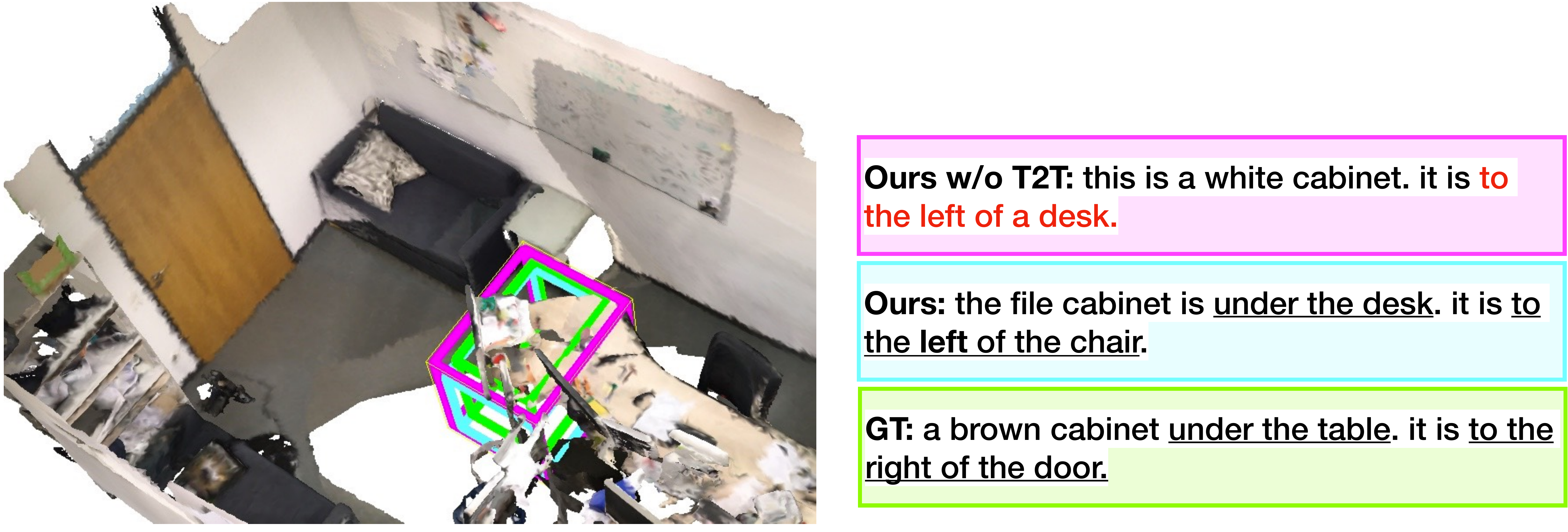}\label{fig:q1}}
\subfigure[]{\includegraphics[width=0.6\textwidth]{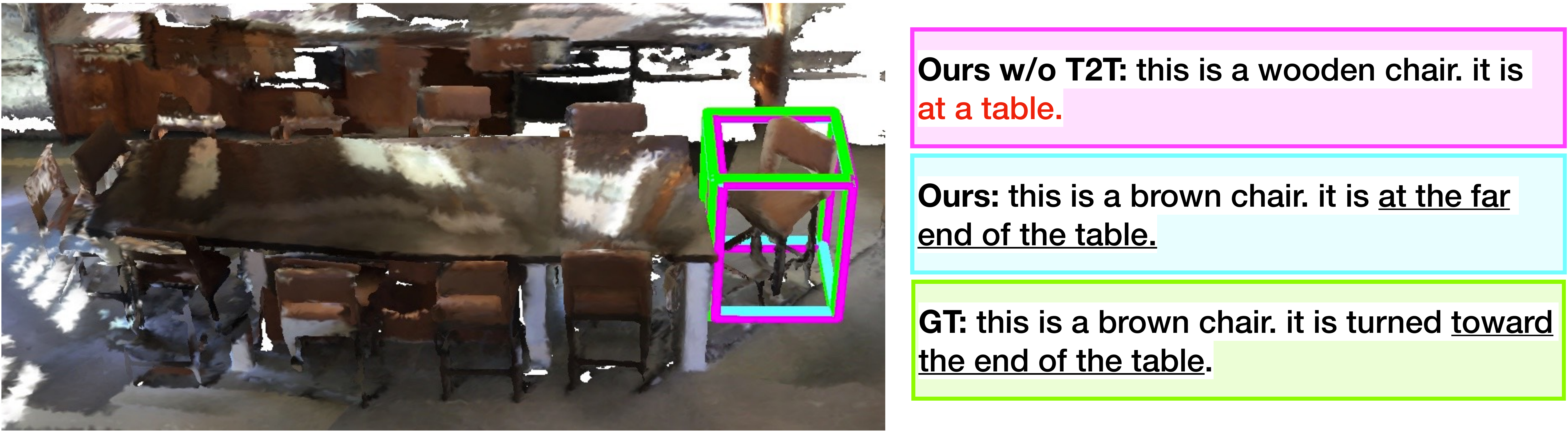}\label{fig:q2}}
\subfigure[]{\includegraphics[width=0.6\textwidth]{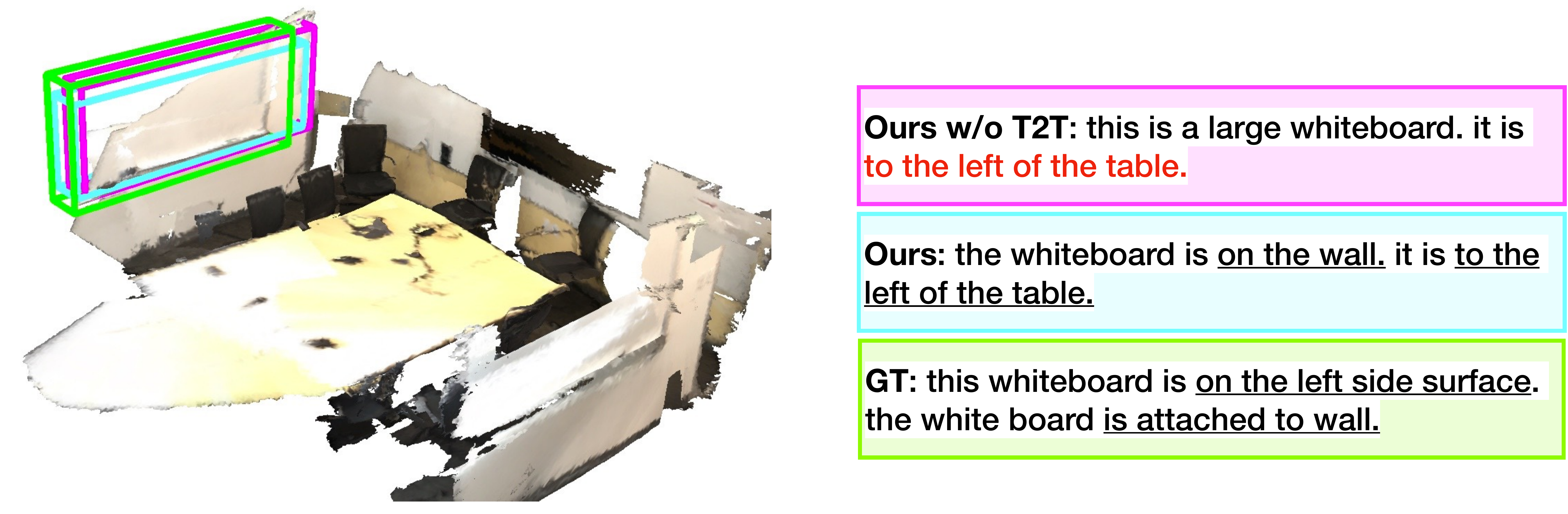}\label{fig:q3}}
\subfigure[]{\includegraphics[width=0.6\textwidth]{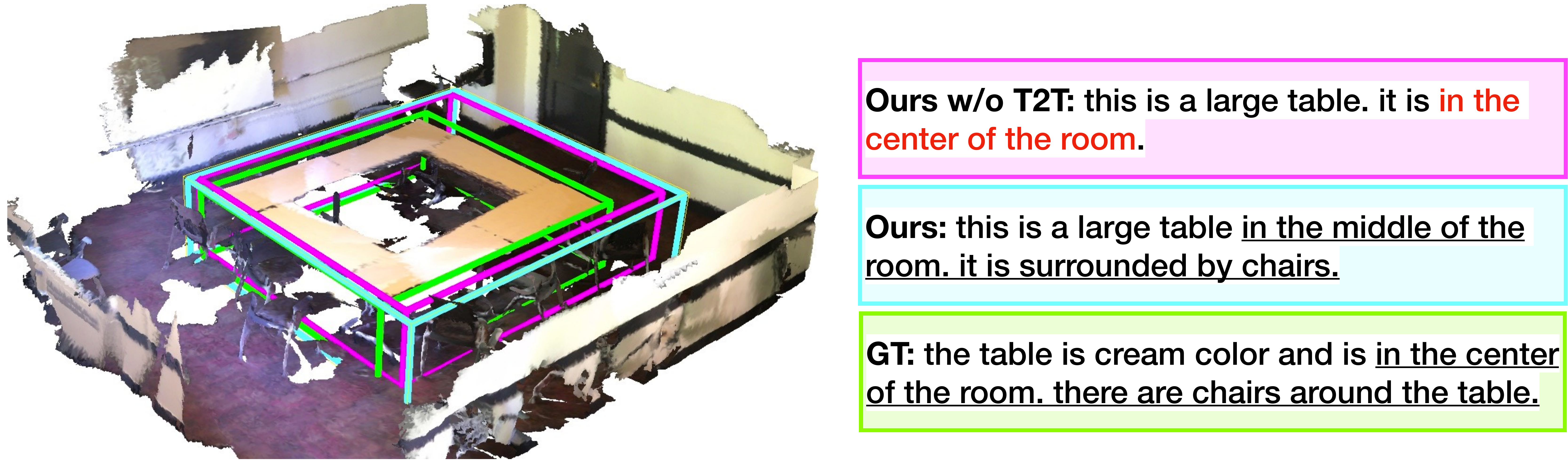}\label{fig:q4}}
\subfigure[]{\includegraphics[width=0.6\textwidth]{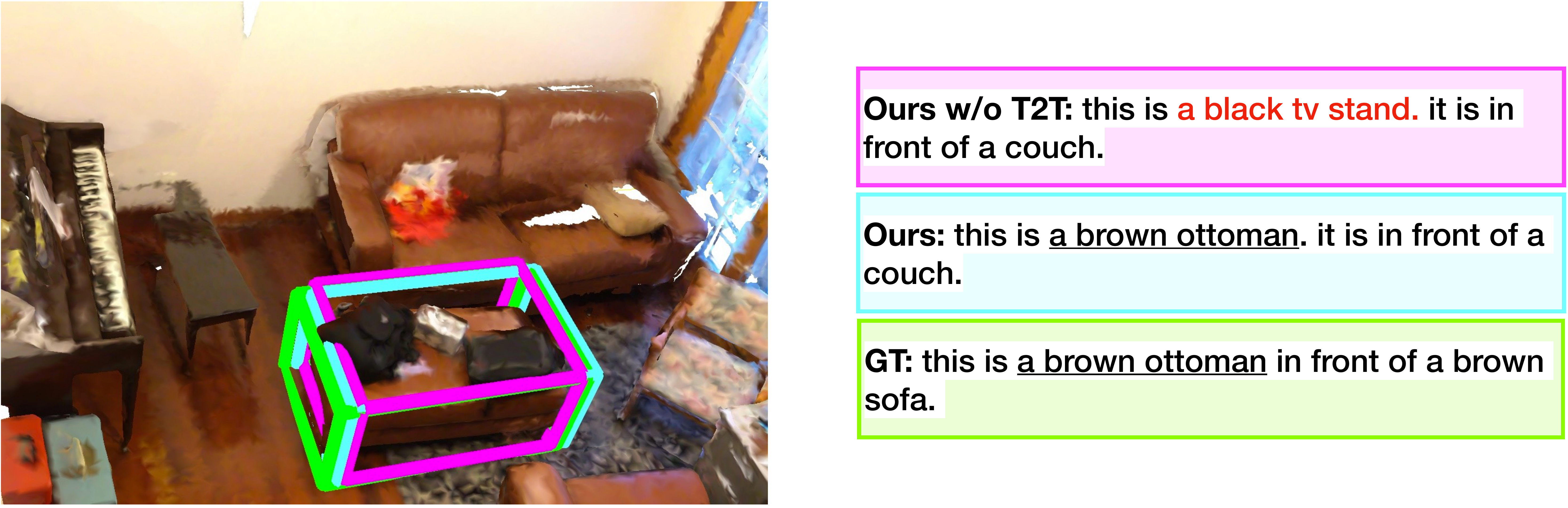}\label{fig:q5}}
\caption{More qualitative results from our methods with and without token-to-token (T2T) spatial relation guidance. Caption boxes share the same color with detection bounding boxes for ground truth (green), ours with T2T (blue), and ours without T2T (pink). Imprecise parts of sentences produced by ours without T2T are marked in red, and correctly expressed descriptions predicted by T2T-guided method are highlighted using underscores.}
\label{fig:more_qualitivie}
\end{figure*}

We display more detect-and-describe results from models trained with and without our proposed token-to-token (T2T) spatial relation guidance in Figure~\ref{fig:more_qualitivie}. In Figure~\ref{fig:q1}, our method with T2T guidance captures two spatial relations for the target object file cabinet, one is with the chair on the right and the other is with the desk on the top. Without T2T guidance, the predicted relation \textit{``to the left of a desk"} is incorrect. Figure~\ref{fig:q2} shows the case when T2T guidance boosts more precise description generation - not merely \textit{``at a table"} but \textit{``at the far end of the table"}. Figure~\ref{fig:q3} and~\ref{fig:q4} present more cases when T2T guidance improves relation variety - relations between whiteboard with wall and table, and between table with the whole room and the surrounding chairs, respectively. Figure~\ref{fig:q5} highlights the case when the lack of T2T guidance could lead to the wrong prediction of the target object itself.

 % -------- figure: positive condition ---------

 % -------- figure: positive condition ---------
% -------- figure: positive condition ---------

\end{appendices}

\end{document}